\documentclass[11pt]{article}

\usepackage[preprint]{acl}
\usepackage{microtype}
\usepackage{graphicx}
\usepackage{alltt}
\usepackage{float}
\usepackage{subfigure}
\usepackage{booktabs} 
\usepackage{listings}
\usepackage{xcolor}
\usepackage{textcomp}
\usepackage[ruled,vlined]{algorithm2e}
\usepackage[utf8]{inputenc}
\usepackage{listings}
\lstdefinelanguage{lean}{}
\usepackage{multirow}
\usepackage{mdframed}
\usepackage{algpseudocode}
\usepackage{bbm}
\usepackage[utf8]{inputenc}
\usepackage{epstopdf}
\usepackage{amsthm}  
\usepackage{amsmath} 
\usepackage{amssymb} 
\usepackage[utf8]{inputenc}

\theoremstyle{plain}

\theoremstyle{definition}

\theoremstyle{remark}

\usepackage{times}
\usepackage{latexsym}

\usepackage[T1]{fontenc}

\usepackage[utf8]{inputenc}

\usepackage{microtype}

\usepackage{inconsolata}

\usepackage{graphicx}

\title{Hierarchical Attention Generates Better Proofs}

\author{
 \textbf{Jianlong Chen\textsuperscript{1,2}},
 \textbf{Chao Li\textsuperscript{2}},
 \textbf{Yang Yuan\textsuperscript{2,3}\footnotemark[2]},
 \textbf{Andrew C Yao\textsuperscript{2,3}\footnotemark[2]}
\\
\\
 \textsuperscript{1}The Chinese University of Hong Kong, Shenzhen \\
 \textsuperscript{2}Shanghai Qi Zhi Institute
 \textsuperscript{3}IIIS, Tsinghua University
\\
 \small{
   \{jianlongchen\}@link.cuhk.edu.cn, \{yuanyang,andrewcyao\}@tsinghua.edu.cn
 }
}

\begin{document}

\maketitle

\renewcommand{\thefootnote}{\fnsymbol{footnote}} 
\footnotetext[2] {Corresponding author.}
\begin{abstract}
Large language models (LLMs) have shown promise in formal theorem proving, but their token-level processing often fails to capture the inherent hierarchical nature of mathematical proofs. We introduce \textbf{Hierarchical Attention}, a regularization method that aligns LLMs' attention mechanisms with mathematical reasoning structures. Our approach establishes a five-level hierarchy from foundational elements to high-level concepts, ensuring structured information flow in proof generation. Experiments demonstrate that our method improves proof success rates by 2.05\% on miniF2F and 1.69\% on ProofNet while reducing proof complexity by 23.81\% and 16.50\% respectively. The code is available at \url{https://github.com/Car-pe/HAGBP}.
\end{abstract}

\section{Introduction}

The intersection of AI and mathematics has emerged as an important research direction in recent years, particularly in the domain of formal theorem proving. 
Proof assistants, such as Lean \cite{de2015lean, moura2021lean}, Coq \cite{coq}, and Isabelle \cite{paulson1994isabelle}, have become key platforms to explore this direction. 
Traditionally, theorem provers primarily rely on search-based methods to systematically explore proof spaces, often guided by complex rule-based techniques or symbolic heuristics \cite{han2021proof, jiang2021lisa, polu2020generative, polu2022formal, lample2022hypertree, jiang2022thor, yang2024leandojo}. 

The advent of large language models (LLMs) has brought a transformative shift, leveraging their capacity for deep contextual understanding to reason about mathematical proofs  \cite{xin2024deepseek, welleck2023llmstep, zhao2023decomposing, jiang2023multilingual, wang2023lego, first2023baldur}. These models excel at generating proofs and tackling a broad array of problems, significantly reducing the need for manually crafted heuristics. However, they still struggle with key challenges in formal theorem proving, often failing to generate difficult proofs or producing unnecessarily long ones.

These limitations arise because mathematics is inherently formal and rigorous, whereas LLMs are primarily designed to process plain token sequences, without explicit formal semantics. Therefore, 
the structured nature of formal concepts --- where dependencies and relationships between concepts play a critical role --- is difficult for LLMs to fully capture. This raises a natural question: 

\begin{center}
	\fbox{
		\parbox{2.4in}{
			\textbf{How to understand structure better?
	}}}
\end{center}

\begin{figure*}[t]
\centering
\includegraphics[width=0.95\linewidth]{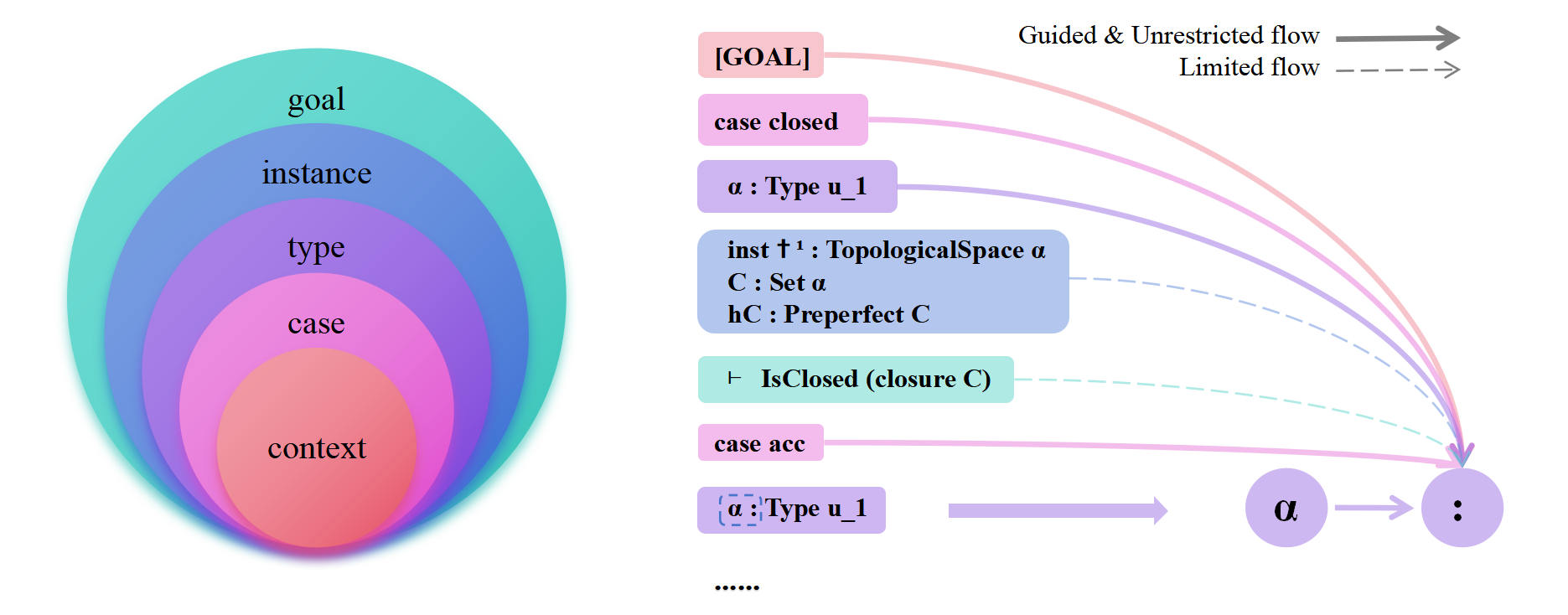}
\caption{Overview of our hierarchical attention framework. \textbf{Left:} The five-level hierarchy from inner (context) to outer (goal) layer, illustrating the natural information flow in mathematical reasoning. \textbf{Right:} A concrete example showing how different components in a theorem proving state are assigned to hierarchical levels, with guided and unrestricted flow (solid arrows) representing allowed attention paths and limited flow (dashed arrows) representing restricted attention paths.}
\label{fig:overview}
\end{figure*}

Mathematical theorem proving exhibits inherent hierarchical structures in the flow of information between different components. While large language models have shown promising results in this domain, their attention mechanisms often fail to capture these natural hierarchies. We propose a novel framework that guides the model's attention patterns to better align with the hierarchical nature of mathematical reasoning, while maintaining flexibility for complex proof steps.

Our key insight is that mathematical reasoning follows a natural hierarchical structure, with information flowing from foundational elements to higher-level concepts. As shown in Figure~\ref{fig:overview}, we formalize this intuition through a five-level hierarchy and implement it by structured attention patterns. This hierarchical framework not only respects the natural dependencies in mathematical proofs but also provides flexibility in attention distribution, allowing the model to capture both local and cross-level relationships necessary for complex reasoning.

Based on this framework, we propose \textbf{Hierarchical Attention}, a novel regularization method aimed at improving structural learning in LLMs. Our approach constructs a hierarchical tree from the input token sequence, assigning levels to tokens and guiding information flow based on these levels. Specifically, we enforce the following constraints: \begin{itemize} 
\item Tokens at higher levels can access information from the same level or lower levels. 
\item Tokens at lower levels are restricted from accessing higher-level information. 
\end{itemize}

Through extensive experiments on multiple theorem-proving benchmarks—including miniF2F~\cite{zheng2021minif2f} and ProofNet~\cite{azerbayev2023proofnet}—our method demonstrates significant improvements in both \textbf{proof success rates} and \textbf{proof conciseness}. Specifically, our approach achieves a 2.05\% improvement in proof success rates while reducing the proof length by 23.81\% in successful cases. These results highlight the advantages of preserving semantic and hierarchical structures in theorem proving. This is further confirmed by our ablation studies and attention pattern analysis.

The main contributions of this work are as follows:

\begin{itemize}
    \item We analyzed the hierarchical structure in mathematical reasoning, from foundational definitions to final goals.
    \item We proposed a new algorithm for better structure learning for LLMs. 
    \item We demonstrated substantial improvements on multiple standard benchmarks in proof accuracy and proof conciseness.
\end{itemize}

\section{Related Work}\label{sec:related_work} 

\textbf{Formal Theorem Proving}. 
Formal theorem proving systems are typically classified into two categories: Automated Theorem Proving (ATP) and Interactive Theorem Proving (ITP). ATP systems aim to discover proofs without human intervention automatically. Saturation-based provers like E~\cite{schulz2002brainiac} and Vampire~\cite{kovacs2013first} use resolution calculus, while specialized solvers like SAT and SMT solvers (e.g., MiniSat~\cite{een2003extensible}, Z3~\cite{de2008z3}) focus on boolean satisfiability and other mathematical theories. Domain-specific systems like GEX~\cite{chou2000deductive} handle geometric problems through specialized deduction rules.

In contrast, ITP systems like Lean~\cite{de2015lean, moura2021lean}, Coq~\cite{coq}, and Isabelle~\cite{paulson1994isabelle} emphasize human-machine collaboration. These systems provide expressive proof languages and sound kernels, enabling mathematicians to formalize theorems and construct proofs in a manner that mirrors informal mathematical reasoning while ensuring logical correctness.

\textbf{Neural Theorem Proving}. 
Neural Theorem Proving has risen to prominence alongside the rapid development of LLMs and more specialized neural architectures for formal reasoning. A central focus has been autoformalization \cite{wang2018first, wang2020exploration, wu2022autoformalization, murphy2024autoformalizing, jiang2022draft, jiang2023multilingual, lu2024process, ying2024lean, azerbayev2023proofnet, liu2023fimo, xin2024deepseek，li2025formalizationalgorithmsoptimizationblock}, which converts informal mathematical statements and proofs into machine-verifiable languages despite the ongoing challenges in semantic alignment. Another key area is premise selection \cite{irving2016deepmath, kucik2018premise, piotrowski2020stateful, ferreira2020natural, ferreira2020premise, wu2022formal, mikula2023magnushammer, holden2025graph}, where models retrieve the most relevant lemmas from vast libraries to aid in proving a target statement. Researchers also tackle proof-step generation \cite{huang2018gamepad, yang2024leandojo, welleck2023llmstep, sanchez2020generating, sanchez2023passport, yang2019learning, polu2020generative, han2021proof, wang2023dt, wang2024proving, lin2024lean, wu2024lean, rute2024graph2tac, dong2025stpselfplayllmtheorem, lin2025goedelproverfrontiermodelopensource, wang2025malotmultiagentleanbasedlong, wang2025kiminaproverpreviewlargeformal, zhang2025leanabellproverposttrainingscalingformal}, aiming to accurately predict the next formal step or tactic, often through auto-regressive models that learn from existing proofs. A further challenge is proof search \cite{loos2017deep, suda2021improving, aygun2020learning, aygun2022proving, chvalovsky2023guiding, rawson2019neurally, rawson2021lazycop, mckeown2023reinforcement, fokoue2023ensemble, abdelaziz2022learning, crouse2021deep, xin2025bfsproverscalablebestfirsttree}, where deep learning-guided algorithms, sometimes using Monte Carlo Tree Search or reinforcement learning, explore and prune massive proof spaces, balancing correctness with computational efficiency. 

\textbf{Hierarchical Attention Mechanisms for Mathematical Reasoning}.
Mathematical documents typically have an implicit multilevel structure, from foundational definitions to the main theorems. Previous studies have attempted to exploit this hierarchical nature by parsing formulas or proofs into trees or graphs to better represent logical structures \cite{wang2017premise, peng2017tree, paliwal2020graph, rawson2020directed}, or by building dependency graphs over entire libraries to capture relationships between statements and lemmas \cite{ferreira2020premise, bauer2024mlfmf}. These approaches, while promising, often depend on carefully crafted rules or programmatically generated data, lacking mechanisms to ensure that neural models respect the partial orders and compositional dependencies inherent in mathematical logic.

The attention mechanism is central to modern Transformer-based models \cite{vaswani2017attention}. Although studies have explored their use in tasks such as generating math problems or document classification \cite{yang2016hierarchical, wu2022automatic}, there is a gap in leveraging attention-based methods explicitly for mathematical reasoning.

\section{Preliminaries}
\subsection{Hierarchical Structure in Lean}
Lean is a strongly typed language, which allows all tokens to be naturally unfolded across multiple semantic levels. These levels align with various components of reasoning, with each successive level built upon the foundations of the preceding ones. The categorization of these layers can be delineated as follows:

\begin{description} 
\item[Lowest or contextual layer:] Contains background information, auxiliary concepts, or general knowledge relevant to the proof ($T_0$: context). 
\item[Intermediate layers:] Include pattern matching and case analysis ($T_1$: case), type declarations and definitions ($T_2$: type), instance declarations and concrete examples ($T_3$: instance) that support the proof.
\item[Highest or goal layer:] Represents the core theorem or proposition to be proved ($T_4$: goal), which relies on the information introduced in the lower layers.
\end{description}

These layers follow a natural partial order: $context \prec case \prec type \prec instance \prec goal$. 
Structuring mathematical reasoning within this hierarchy yields two key benefits: 
\begin{itemize} 
\item \emph{Proper Scoping}: Contextual elements and definitions are confined to their appropriate levels. Intuitively, each concept is most meaningfully analyzed in conjunction with others at the same level, ensuring logical coherence and clarity.
\item \emph{Clear Semantic Flow}: The reasoning progresses seamlessly from foundational definitions to the final goal, reflecting the natural and intuitive structure of mathematical arguments.
\end{itemize}

\subsection{Information Flow}
We want to exploit the hierarchical structure by incorporating flow control into the model. Let $T$ be the set of all tokens of the input theorem. We use $t_i, t_j$ to denote individual tokens, $L$ for the number of transformer layers, and $1\leq l \leq L $ for layer indices. For tokens $t_i, t_j$ in layer $l$, we define:
\begin{itemize}
    \item $att_l(t_i,t_j)$: attention score from $t_i$ to $t_j$, representing how much $t_i$ will affect embedding of $t_j$ at layer $l$, 
    \item $M_{ij}$: binary attention mask, controlling the information flow from $t_i$ to $t_j$,
    \item $\alpha_l = 1-l/L$: layer-wise adaptation factor, which attenuates flow control for deeper layers.
\end{itemize}

We use $\text{level}(t_i)$ to denote the hierarchical level of token $t_i$, taking value from $\{0,1,2,3,4\}$, corresponding to the five levels in our hierarchy. By controlling attention flow based on these levels, we encourage the model to follow natural mathematical reasoning patterns, where higher-level concepts build upon lower-level foundations.

\section{Approach}
To enhance the model's comprehension of the hierarchical structure and its ability to reason in alignment with it, we propose a two-step approach. First, we extract the flow pattern from the input by identifying different hierarchical levels in mathematical statements. Second, we guide the model's attention through a specialized loss function that encourages the model to respect these hierarchical relationships during training.

\subsection{Extract Flow Pattern}
In mathematical reasoning, different components of a statement naturally form a hierarchy. We identify five distinct levels (labeled 0 to 4): basic tokens, case-specific elements, type definitions, problem instances, and goal statements. The flow from token $t_i$ to token $t_j$ may follow one of three types, based on their hierarchical levels:
\begin{equation}
\begin{cases}
    \text{Unrestricted} & \text{if } \text{level}(t_i) = \text{level}(t_j) \\
    \text{Guided} & \text{if } \text{level}(t_i) < \text{level}(t_j) \\
    \text{Limited} & \text{if } \text{level}(t_i) > \text{level}(t_j)
\end{cases}
\label{flow_structure}
\end{equation}
While sharing similar algorithmic treatment, unrestricted and guided flows are distinctly categorized due to their different functional roles. Our experiments \ref{sec:visualization_analysis} show these flows develop distinct patterns during training: unrestricted flow shows reduced attention proportion, while guided flow demonstrates increased dominance. This structure ensures that semantic dependencies respect the hierarchical nature of mathematical reasoning, with tokens primarily attending to those at the same or lower levels, while limiting attention in the reverse direction to maintain logical consistency.

\subsection{Algorithm Implementation}
Based on these flow patterns, we implement a hierarchical attention mechanism as shown in Algorithm~\ref{alg:main}. The algorithm first parses the input into different hierarchical levels using string pattern matching to identify key mathematical components. It then constructs attention masks and computes a flow loss that penalizes attention patterns violating hierarchical constraints.

\begin{algorithm}[t]
\caption{Hierarchical Attention Implementation}
\label{alg:main}
\KwIn{Theorem text $T$, Model layers $L$}
\KwOut{Flow loss $\mathcal{L}_{flow}$}
\BlankLine
\tcc{Initialize hierarchical levels}
Parse input into level sets $\{T_0,...,T_4\}$ \tcp*{Using Algorithm~\ref{alg:parsing}}\;
Initialize attention mask $M$, $\mathcal{L}_{flow} \gets 0$\;
\BlankLine
\For{each layer $l$ in $1$ to $L$}{
    $\alpha_l \gets (1 - l/L)$ \tcp*{Layer adaptation factor}
    \For{tokens $t_i, t_j$ in input}{
        \tcc{Construct attention mask}
        \If{$\text{level}(t_i) \leq \text{level}(t_j)$}{
            $M_{ij} \gets 1$ \tcp*{Allow upward/horizontal flow}
        }\Else{
            $M_{ij} \gets 0$ \tcp*{Limit downward flow}
        }
        
        \tcc{Compute loss contribution}
        $invalid_{flow} \gets att_l(t_i,t_j) \cdot (1 - M_{ij})$\;
        $\mathcal{L}_{flow} \gets \mathcal{L}_{flow} + \alpha_l \cdot \text{ReLU}(invalid_{flow})$\;
    }
}
$\mathcal{L}_{flow} \gets \mathcal{L}_{flow}/|T|$\;
\Return{$\mathcal{L}_{flow}$}\;
\end{algorithm}

The flow loss $\mathcal{L}_{flow}$ penalizes attention patterns that violate hierarchical constraints:
\begin{equation}
\mathcal{L}_{flow} = \frac{1}{|T|}\sum_{l=1}^L \alpha_l \cdot \sum_{i,j} \text{ReLU}(att_l(t_i,t_j) \cdot (1-M_{ij}))
\end{equation}
where $\alpha_l = (1-\frac{l}{L})$ provides stronger regularization in earlier layers while allowing more flexibility in later layers.

The final training objective combines this flow loss with the standard cross-entropy loss $\mathcal{L}_{LM}$:
\begin{equation}
\mathcal{L} = \mathcal{L}_{LM} + \lambda \mathcal{L}_{flow}
\end{equation}
where $\lambda$ controls the strength of hierarchical constraints. A larger $\lambda$ enforces stricter adherence to the hierarchy, while a smaller value allows more flexible attention patterns.

In summary, our approach:
\begin{itemize}
    \item Identifies natural hierarchical levels in mathematical statements.
    \item Guides attention patterns to respect hierarchical relationships.
    \item Enables flexible reasoning through layer-wise adaptation.
\end{itemize}




\section{Experiments}
In this section, we evaluate our method through comprehensive experiments on multiple theorem-proving benchmarks. 


\subsection{Experimental Setup}
\paragraph{Training Data and Configuration}
We use LeanDojo Benchmark 4\footnote{Yang, K. (2023). LeanDojo Benchmark (v1) [Data set]. Zenodo. \url{https://doi.org/10.5281/zenodo.8016386}} as our training dataset. The training process involves fine-tuning a Pythia-2.8B\footnote{\url{https://huggingface.co/EleutherAI/pythia-2.8b}}~\cite{biderman2023pythia} model for 3 epochs. Detailed hyperparameters and training configurations are provided in Appendix~\ref{app:training_details}.

\paragraph{Evaluation Protocol}
We conduct comprehensive evaluations across four benchmark datasets: miniF2F (test/valid)\footnote{\url{https://huggingface.co/datasets/cat-searcher/minif2f-lean4}} and ProofNet (test/valid)\footnote{\url{https://huggingface.co/datasets/UDACA/proofnet-lean4}}. Our evaluation employs two complementary strategies: best-first search and single-pass sampling, to demonstrate the robustness of our method (detailed algorithms in Appendix~\ref{app:evaluation_algorithm}).

For both strategies, we define the computation budget as $K \times T$, where $T$ indicates the number of expansion iterations, which is set to 100 across all our experiments, and $K = N \times S$. For the best-first search, N represents the number of parallel search attempts and S denotes the number of tactics generated per expansion. For single-pass sampling, N represents the total number of sampling attempts per problem, while S is fixed to 1 as only one tactic is attempted at each expanded node. The search process employs parallel sampling with fixed time constraints per theorem. In the following sections, we use $K$ to denote the product of N and S for simplicity.

Our method is a general-purpose fine-tuning technique that can be applied to any formal theorem-proving system. For empirical validation, we chose LLMSTEP~\cite{welleck2023llmstep} as our primary baseline, which provides full access to its model, dataset, and hyperparameters, ensuring complete reproducibility of our comparative analysis.

\subsection{Main Results}
We present a comparative analysis of our method against the baseline, highlighting its performance and advancements.

\begin{figure*}[t]
\centering
\includegraphics[width=0.48\linewidth]{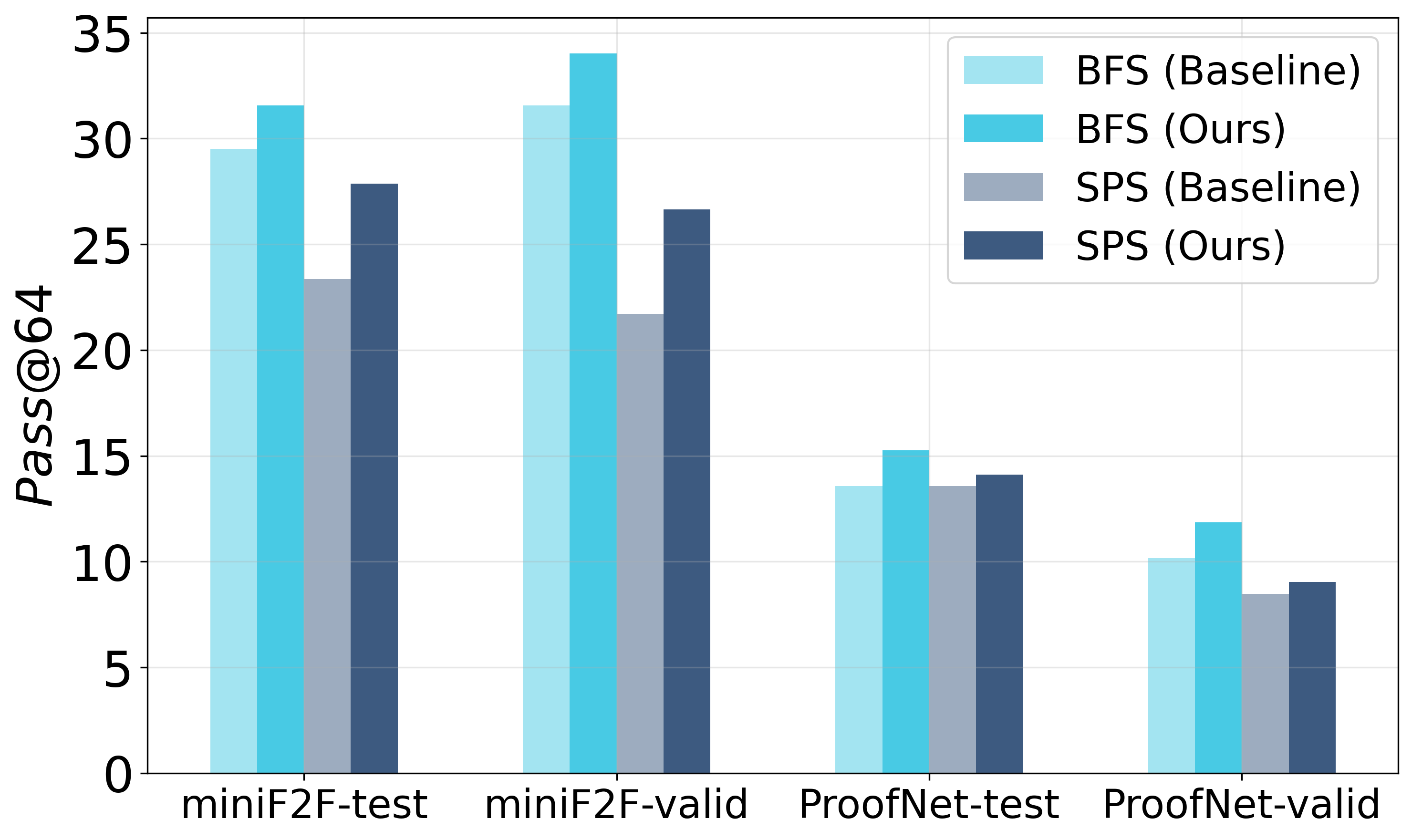}\hfill
\includegraphics[width=0.48\linewidth]{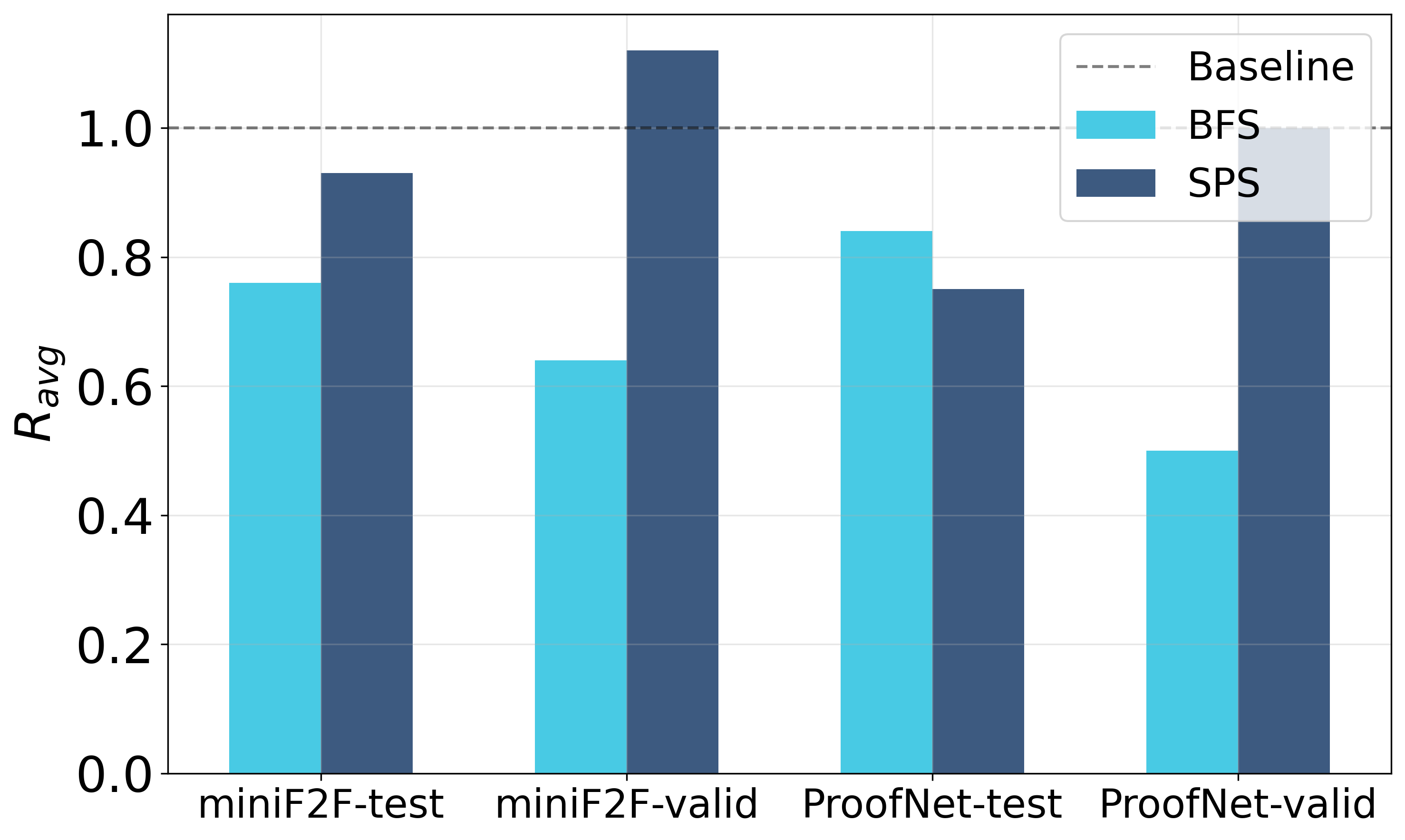}
\caption{Performance comparison between our method and baseline at $K = 64$. \textbf{Left:} Pass rate comparison across miniF2F (test/valid) and ProofNet (test/valid) datasets. Best-first search (BFS) consistently outperforms single-pass sampling (SPS), with our method further enhancing BFS performance. Solid bars represent our method while transparent bars represent the baseline. \textbf{Right:} Proof complexity ratio ($R_{avg}$), where values below 1.0 (dashed line) indicate more concise proofs. Our method with BFS achieves consistent complexity reductions across all datasets.}
\label{fig:main_results}
\end{figure*}

\paragraph{Metrics}
We evaluate our method using two key metrics: pass@$K$ accuracy and proof complexity. The pass@$K$ metric measures the model's ability to generate a valid proof within $K$ sampling attempts, where $K = N \times S$ represents the total number of tactic samples considered during this iteration of proof search.

For proof conciseness analysis, we measure the number of proof steps required to solve the goals. Let $\mathcal{T}_{com}$ be the set of theorems successfully proved by both methods with different proof lengths. For each theorem $t \in \mathcal{T}_{com}$, we define its proof complexity as:

\begin{equation}
C(t, m) = |p_{t,m}|
\end{equation}

where $p_{t,m}$ is the proof generated for theorem $t$ using method $m$, and $|p_{t,m}|$ denotes the number of proof steps. We then compute the average complexity ratio:

\begin{equation}
R_{avg} = \frac{1}{|\mathcal{T}_{com}|} \sum_{t \in \mathcal{T}_{com}} \frac{C(t, \text{ours})}{C(t, \text{baseline})}
\end{equation}

This metric provides a direct measure of our method's proof conciseness, where $R_{avg} < 1$ indicates that our method generally produces shorter proofs. Note that we only consider theorems where \textbf{both methods succeed but generate proofs of different lengths}, as this provides a meaningful comparison of the proof conciseness. We also report $\text{Diff.}$ (\%), which indicates the percentage of such theorems among all theorems that both methods successfully prove, reflecting how often the methods differ in their proof strategies.

\begin{table}[t]
\begin{center}
\caption{Results on miniF2F test set with best-first search strategy.}
\label{tab:minif2f_test_search}
\scalebox{0.75}{
\begin{tabular}{lcccccc}
\toprule
\multirow{2}{*}{$K$} & \multicolumn{2}{c}{PASS(\%)} & \multicolumn{2}{c}{COMPLEXITY} & \multirow{2}{*}{$R_{avg}$} & \multirow{2}{*}{Diff. (\%)} \\
\cmidrule(lr){2-3} \cmidrule(lr){4-5}
& Baseline & Ours & Baseline & Ours & & \\
\midrule
1 & \textbf{14.75} & 14.34 & - & - & - & - \\
2 & \textbf{18.44} & 17.62 & - & - & - & - \\
4 & 22.54 & \textbf{23.36} & - & - & - & - \\
8 & 26.23 & 26.23 & 2.00 & \textbf{1.86} & \textbf{0.93} & 11.67 \\
16 & \textbf{29.10} & 28.28 & 2.11 & \textbf{1.50} & \textbf{0.71} & 13.24 \\
32 & 29.51 & \textbf{31.15} & 1.89 & \textbf{1.67} & \textbf{0.88} & 12.50 \\
64 & 29.51 & \textbf{31.56} & 2.10 & \textbf{1.60} & \textbf{0.76} & 8.11 \\
\bottomrule
\end{tabular}
}
\end{center}
\end{table}

\begin{table}[t]
\begin{center}
\caption{Results on miniF2F validation set with best-first search strategy.}
\label{tab:minif2f_valid_search}
\scalebox{0.75}{
\begin{tabular}{lcccccc}
\toprule
\multirow{2}{*}{$K$} & \multicolumn{2}{c}{PASS(\%)} & \multicolumn{2}{c}{COMPLEXITY} & \multirow{2}{*}{$R_{avg}$} & \multirow{2}{*}{$\text{Diff.}$ (\%)} \\
\cmidrule(lr){2-3} \cmidrule(lr){4-5}
& Baseline & Ours & Baseline & Ours & & \\
\midrule
1 & 12.70 & \textbf{13.52} & - & - & - & - \\
2 & \textbf{15.16} & 14.75 & - & - & - & - \\
4 & 20.49 & \textbf{23.77} & - & - & - & - \\
8 & 27.05 & \textbf{29.51} & 2.83 & \textbf{2.67} & \textbf{0.94} & 9.68 \\
16 & 31.15 & \textbf{33.20} & 2.89 & \textbf{1.89} & \textbf{0.65} & 13.89 \\
32 & 31.56 & \textbf{34.02} & 3.11 & \textbf{2.00} & \textbf{0.64} & 12.00 \\
64 & 31.56 & \textbf{34.02} & 3.11 & \textbf{2.00} & \textbf{0.64} & 12.68 \\
\bottomrule
\end{tabular}
}
\end{center}
\end{table}

\begin{table}[t]
\begin{center}
\caption{Results on miniF2F test set with single-pass sampling strategy.}
\label{tab:minif2f_test_sample}
\scalebox{0.75}{
\begin{tabular}{lcccccc}
\toprule
\multirow{2}{*}{$K$} & \multicolumn{2}{c}{PASS(\%)} & \multicolumn{2}{c}{COMPLEXITY} & \multirow{2}{*}{$R_{avg}$} & \multirow{2}{*}{$\text{Diff.}$ (\%)} \\
\cmidrule(lr){2-3} \cmidrule(lr){4-5}
& Baseline & Ours & Baseline & Ours & & \\
\midrule
1 & 9.84 & \textbf{18.44} & - & - & - & - \\
2 & 12.30 & \textbf{20.90} & - & - & - & - \\
4 & 16.80 & \textbf{24.18} & - & - & - & - \\
8 & 19.63 & \textbf{25.00} & 1.95 & \textbf{1.86} & \textbf{0.95} & 51.16 \\
16 & 20.49 & \textbf{26.23} & \textbf{1.85} & 1.92 & 1.04 & 26.00 \\
32 & 23.36 & \textbf{26.64} & 1.83 & \textbf{1.78} & \textbf{0.97} & 15.38 \\
64 & 23.36 & \textbf{27.87} & 2.00 & \textbf{1.85} & \textbf{0.93} & 23.21 \\
\bottomrule
\end{tabular}
}
\end{center}
\end{table}

\begin{table}[t]
\begin{center}
\caption{Results on miniF2F validation set with single-pass sampling strategy.}
\label{tab:minif2f_valid_sample}
\scalebox{0.75}{
\begin{tabular}{lcccccc}
\toprule
\multirow{2}{*}{$K$} & \multicolumn{2}{c}{PASS(\%)} & \multicolumn{2}{c}{COMPLEXITY} & \multirow{2}{*}{$R_{avg}$} & \multirow{2}{*}{$\text{Diff.}$ (\%)} \\
\cmidrule(lr){2-3} \cmidrule(lr){4-5}
& Baseline & Ours & Baseline & Ours & & \\
\midrule
1 & 9.43 & \textbf{14.75} & - & - & - & - \\
2 & 12.30 & \textbf{15.16} & - & - & - & - \\
4 & 16.80 & \textbf{20.08} & - & - & - & - \\
8 & 18.03 & \textbf{21.13} & 2.33 & \textbf{1.73} & \textbf{0.74} & 37.50 \\
16 & 18.44 & \textbf{24.59} & 1.95 & \textbf{1.89} & \textbf{0.97} & 43.18 \\
32 & 20.08 & \textbf{25.41} & 1.92 & 1.92 & 1.00 & 27.66 \\
64 & 21.72 & \textbf{26.64} & \textbf{2.12} & 2.38 & 1.12 & 16.33 \\
\bottomrule
\end{tabular}
}
\end{center}
\end{table}

\paragraph{Overview}
Figure~\ref{fig:main_results} presents a comprehensive evaluation of our method across miniF2F (test/valid) and ProofNet (test/valid) datasets at $K = 64$. The results demonstrate that best-first search (BFS) is the superior search strategy across all datasets, consistently outperforming single-pass sampling (SPS). When combined with our hierarchical attention mechanism, BFS achieves even stronger results. For example, on the miniF2F test set, our method improves the pass rate by 2.05\% while reducing proof complexity by 23.81\%. Similar improvements are observed on the ProofNet test set, with a 1.69\% increase in pass rate and a 16.50\% reduction in proof complexity. Notably, our method also significantly improves SPS performance, particularly on the miniF2F dataset where we observe pass rate improvements of 4.51\% and 4.92\% on test and valid sets respectively. 

\paragraph{Results on miniF2F}
Tables~\ref{tab:minif2f_test_search}-\ref{tab:minif2f_valid_sample} present comprehensive results on the miniF2F benchmark. With best-first search, our method achieves consistent improvements in pass rates at higher computation budgets, reaching 31.56\% on test set (vs. baseline's 29.51\%) and 34.02\% on validation set (vs. baseline's 31.56\%). The performance gain becomes more pronounced as the computation budget increases, particularly when $K$ exceeds 16.

Single-pass sampling results also demonstrate the effectiveness of our method, achieving 27.87\% and 26.64\% pass rates on test and validation sets respectively at $K$ = 64, compared to baseline's 23.36\% and 21.72\%. This represents substantial improvements of 4.51\% and 4.92\% respectively.

For proof conciseness evaluation, we focus on higher computation budget scenarios ($K$ $\geq$ 8) where sufficient successful proofs are available for reliable complexity comparison. At $K$ = 64, our method demonstrates significant advantages in proof conciseness with the search strategy, reducing the average proof length from 3.11 to 2.00 steps ($R_{avg}$ = 0.64) on the validation set and from 2.10 to 1.60 steps ($R_{avg}$ = 0.76) on the test set. The reliability of these complexity metrics is supported by a substantial proportion of comparable cases ($\text{Diff.}$), where both methods succeed but with different proof lengths. For instance, at $K$ = 64 with best-first search, these comparable cases constitute 8.11\% and 12.68\% of all successful proofs for test and validation sets respectively, providing a meaningful sample size for complexity comparison. Similar reliability is observed in single-pass sampling, where $\text{Diff.}$ reaches 23.21\% and 16.33\%, ensuring the robustness of the reported complexity improvements.

\paragraph{Results on ProofNet}

\begin{table}[t]
\begin{center}
\caption{Results on ProofNet test set with best-first search strategy.}
\label{tab:proofnet_test_search}
\scalebox{0.75}{
\begin{tabular}{lcccccc}
\toprule
\multirow{2}{*}{$K$} & \multicolumn{2}{c}{PASS(\%)} & \multicolumn{2}{c}{COMPLEXITY} & \multirow{2}{*}{$R_{avg}$} & \multirow{2}{*}{$\text{Diff.}$ (\%)} \\
\cmidrule(lr){2-3} \cmidrule(lr){4-5}
& Baseline & Ours & Baseline & Ours & & \\
\midrule
16 & 11.86 & 11.86 & - & - & - & - \\
32 & 13.56 & \textbf{14.69} & 1.83 & 1.83 & 1.00 & 28.57 \\
64 & 13.56 & \textbf{15.25} & 2.00 & \textbf{1.67} & \textbf{0.84} & 26.09 \\
\bottomrule
\end{tabular}
}
\end{center}
\end{table}

\begin{table}[t]
\begin{center}
\caption{Results on ProofNet validation set with best-first search strategy.}
\label{tab:proofnet_valid_search}
\scalebox{0.75}{
\begin{tabular}{lcccccc}
\toprule
\multirow{2}{*}{$K$} & \multicolumn{2}{c}{PASS(\%)} & \multicolumn{2}{c}{COMPLEXITY} & \multirow{2}{*}{$R_{avg}$} & \multirow{2}{*}{$\text{Diff.}$ (\%)} \\
\cmidrule(lr){2-3} \cmidrule(lr){4-5}
& Baseline & Ours & Baseline & Ours & & \\
\midrule
16 & 9.04 & \textbf{10.73} & - & - & - & - \\
32 & 9.04 & \textbf{10.73} & 2.00 & \textbf{1.00} & \textbf{0.50} & 12.50 \\
64 & 10.17 & \textbf{11.86} & 2.00 & \textbf{1.00} & \textbf{0.50} & 18.75 \\
\bottomrule
\end{tabular}
}
\end{center}
\end{table}

\begin{table}[t]
\begin{center}
\caption{Results on ProofNet test set with single-pass sampling strategy.}
\label{tab:proofnet_test_sample}
\scalebox{0.75}{
\begin{tabular}{lcccccc}
\toprule
\multirow{2}{*}{$K$} & \multicolumn{2}{c}{PASS(\%)} & \multicolumn{2}{c}{COMPLEXITY} & \multirow{2}{*}{$R_{avg}$} & \multirow{2}{*}{$\text{Diff.}$ (\%)} \\
\cmidrule(lr){2-3} \cmidrule(lr){4-5}
& Baseline & Ours & Baseline & Ours & & \\
\midrule
16 & 9.60 & \textbf{11.30} & - & - & - & - \\
32 & 10.17 & \textbf{12.80} & 2.00 & 2.00 & 1.00 & 31.25 \\
64 & 13.56 & \textbf{14.12} & 2.40 & \textbf{1.80} & \textbf{0.75} & 21.74 \\
\bottomrule
\end{tabular}
}
\end{center}
\end{table}

\begin{table}[t]
\begin{center}
\caption{Results on ProofNet validation set with single-pass sampling strategy.}
\label{tab:proofnet_valid_sample}
\scalebox{0.75}{
\begin{tabular}{lcccccc}
\toprule
\multirow{2}{*}{$K$} & \multicolumn{2}{c}{PASS(\%)} & \multicolumn{2}{c}{COMPLEXITY} & \multirow{2}{*}{$R_{avg}$} & \multirow{2}{*}{$\text{Diff.}$ (\%)} \\
\cmidrule(lr){2-3} \cmidrule(lr){4-5}
& Baseline & Ours & Baseline & Ours & & \\
\midrule
16 & 7.34 & 7.34 & - & - & - & - \\
32 & \textbf{8.47} & 7.34 & 1.50 & 1.50 & 1.00 & 18.18 \\
64 & 8.47 & \textbf{9.04} & 1.50 & 1.50 & 1.00 & 30.77 \\
\bottomrule
\end{tabular}
}
\end{center}
\end{table}

\begin{figure*}[t]
\centering
\includegraphics[width=0.85\linewidth]{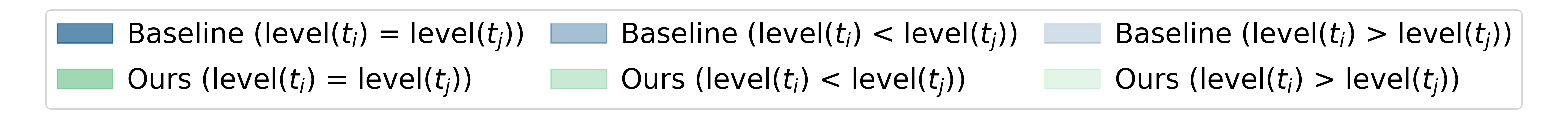}
\includegraphics[width=0.48\linewidth]{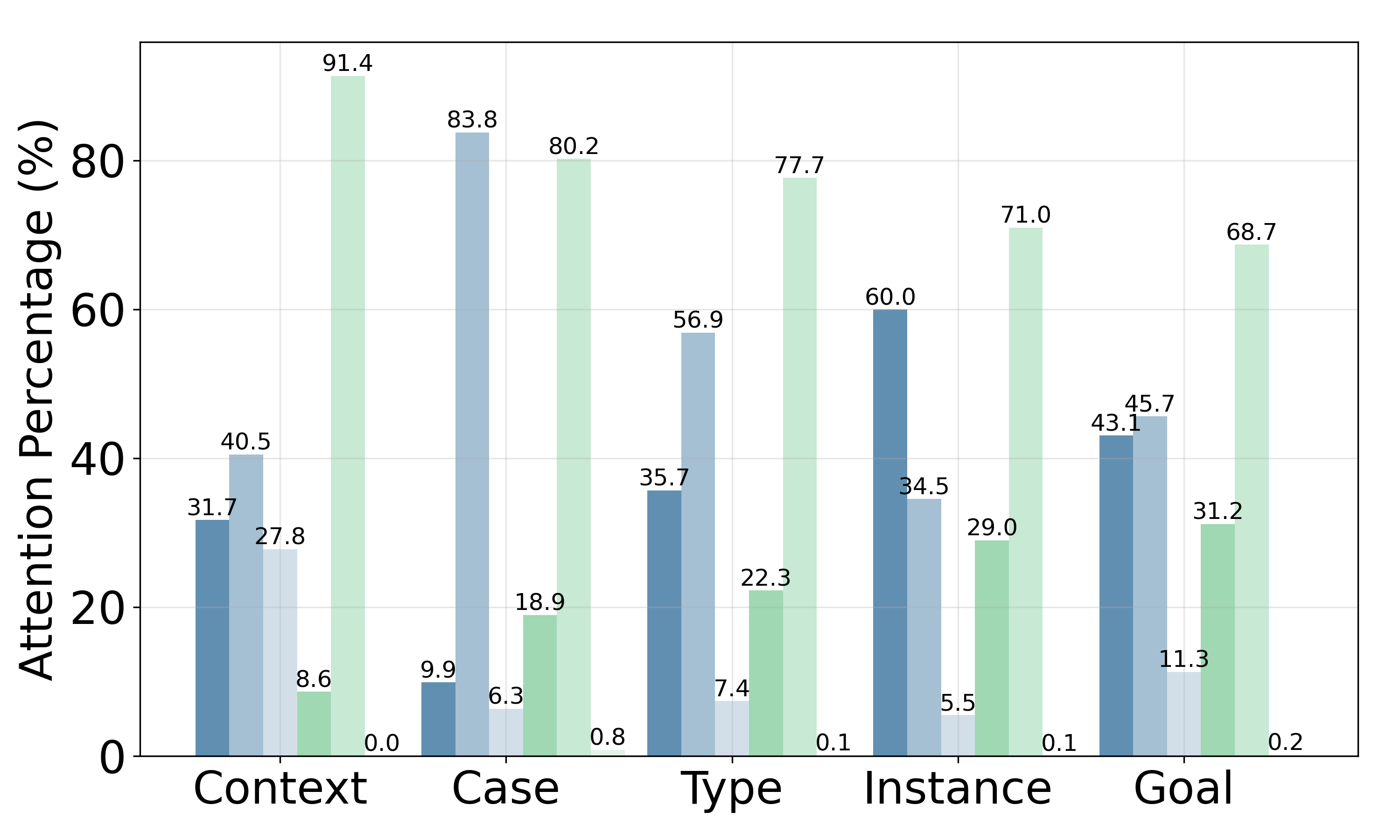}\hfill
\includegraphics[width=0.48\linewidth]{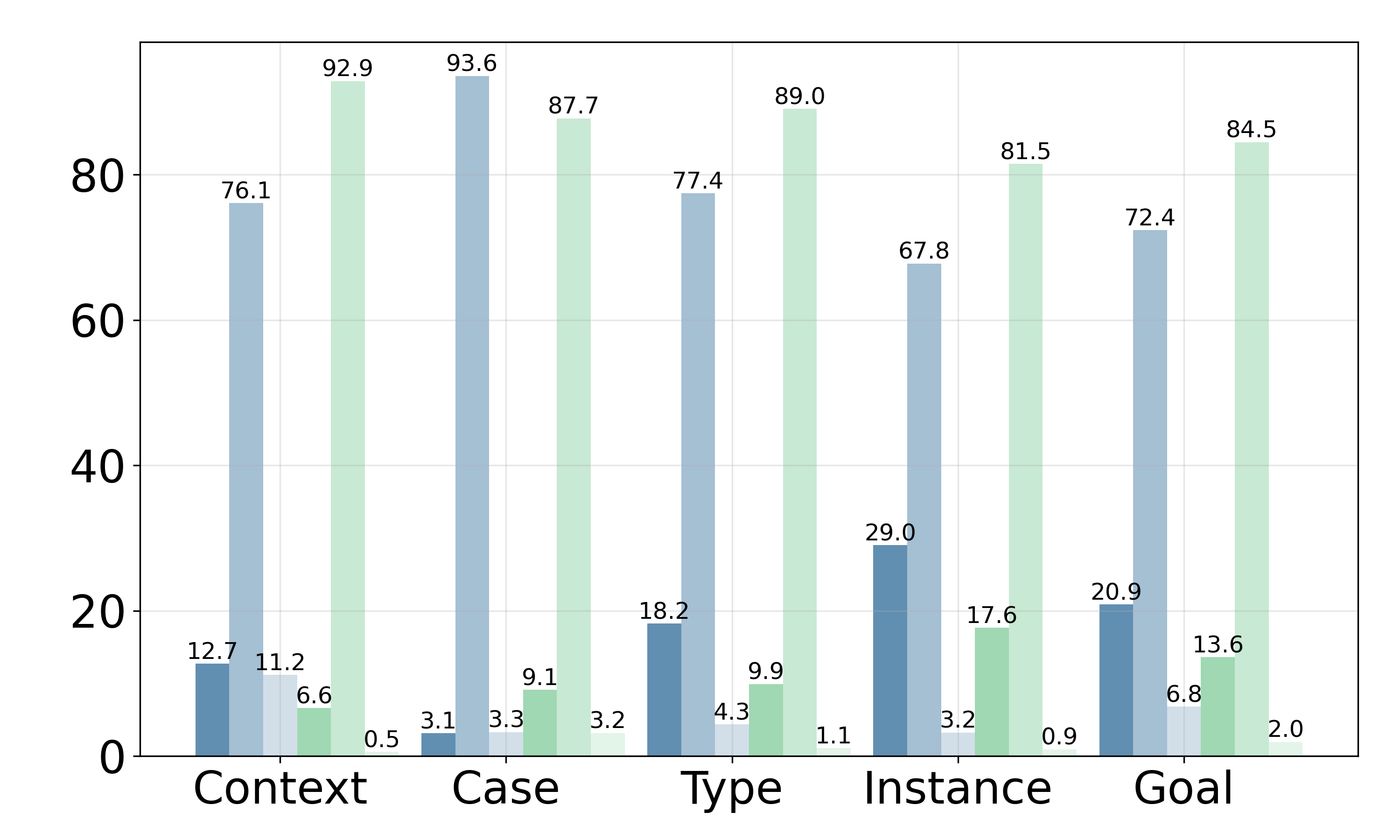}
\caption{Attention distribution analysis in different layers. \textbf{Left:} Hierarchy-constrained layers (where $\alpha_l \neq 0$). \textbf{Right:} Unconstrained layers (where $\alpha_l = 0$). This visualization is derived from averaging attention patterns across all evaluation samples on the LeanDojo Benchmark 4 test set. The x-axis represents different hierarchical levels, while the y-axis shows the percentage of attention scores, combining both cases where the level's tokens serve as source ($t_i$) and target ($t_j$). Blue and green bars represent the baseline and our method respectively, with different transparency levels indicating different attention flow types based on the relationship between source level($t_i$) and target level($t_j$).}
\label{fig:attention_analysis}
\end{figure*}

Tables~\ref{tab:proofnet_test_search}-\ref{tab:proofnet_valid_sample} present the results on ProofNet benchmark. With best-first search strategy, our method achieves consistent improvements in PASS rates at higher computation budgets, reaching 15.25\% on test set (vs. baseline's 13.56\%) and 11.86\% on validation set (vs. baseline's 10.17\%) at $K = 64$.

Single-pass sampling results also demonstrate the effectiveness of our method. On the test set, our method shows consistent improvements across computation budgets, achieving 14.12\% at $K = 64$ compared to baseline's 13.56\%, with improvements ranging from 1.70\% to 2.63\%. On the validation set, while performance is initially comparable at $K = 16$ (both 7.34\%), our method shows improvement at higher computation budgets, reaching 9.04\% at $K = 64$ compared to baseline's 8.47\%.

For proof conciseness evaluation at $K = 64$, our method demonstrates significant advantages across all settings. With the search strategy, the average proof length decreases from 2.00 to 1.67 steps ($R_{avg} = 0.84$) on the test set and from 2.00 to 1.00 steps ($R_{avg} = 0.50$) on the validation set, based on 26.09\% and 18.75\% of differing proofs respectively. The single-pass sampling shows similar improvements with $R_{avg} = 0.75$ on the test set across 21.74\% of differing cases.



\subsection{Visualization and Analysis of Attention Patterns}
\label{sec:visualization_analysis}

The attention distribution analysis shown in Figure~\ref{fig:attention_analysis} demonstrates that our mechanism successfully implements and maintains the designed information flow structure (Equation~\ref{flow_structure}) throughout the model. Our analysis reveals several key findings across both constrained and unconstrained layers:

\subsubsection{Implementation of Limited Flow Constraint}
Our approach enforces the limited flow constraint by minimizing attention flows from higher to lower levels across all layers. In constrained layers (Figure~\ref{fig:attention_analysis}, left), this is evidenced by the near-zero percentages of $\text{level}(t_i) > \text{level}(t_j)$ attention across all hierarchical levels, compared to the baseline's substantial invalid flows ranging from 5.5\% to 27.8\%. Remarkably, this pattern persists in unconstrained layers (Figure~\ref{fig:attention_analysis}, right), where invalid flows remain minimal (ranging from 0.5\% to 3.2\% across different levels), demonstrating the robustness of our hierarchical structure.

\subsubsection{Effectiveness of Guided Flow Design}
Our method successfully implements and maintains the guided flow design throughout the model. In constrained layers, the goal level effectively integrates information from lower levels with 68.7\% upward attention while restricting reverse flows to just 0.2\%. Type and instance levels receive substantial guided information flow from lower levels (77.7\% and 71.0\% respectively), demonstrating strong hierarchical information propagation. This pattern strengthens in unconstrained layers, where the goal level receives even stronger attention from lower levels (84.5\%), and type and instance levels maintain robust upward attention flows (89.0\% and 81.5\% respectively).

\subsubsection{Global Impact on Model Behavior}
The consistency of hierarchical patterns between constrained and unconstrained layers is particularly significant, indicating that our method induces a global, coherent hierarchical information processing framework. Rather than merely responding to external constraints, the model appears to have internalized the hierarchical structure, as evidenced by the preservation of desired attention patterns in unconstrained layers. This seamless continuation of attention patterns throughout the model architecture suggests that our hierarchical attention mechanism effectively shapes the model's overall information processing strategy, establishing a stable and consistent hierarchical flow structure.

\section{Conclusion}

We introduced Hierarchical Attention, a regularization method that aligns transformer attention with mathematical reasoning structures through a five-level hierarchy. Our approach balances structured information flow with the flexibility needed for complex proofs through layer-wise adaptation. Experimental results show improved proof success rates and conciseness across multiple benchmarks, while attention pattern analysis confirms the method's effectiveness in helping models internalize mathematical hierarchies. The consistent improvements demonstrate a promising direction for bridging neural language models and mathematical reasoning.

\section*{Limitations}
Our approach has three main limitations: (1) the hierarchy definition is specific to Lean's semantics and may require adaptation for other proof languages, (2) the fixed hierarchy structure may limit dynamic reasoning patterns, and (3) data constraints prevented evaluation on advanced models like DeepSeek-Prover~\cite{xin2024deepseek} and InternLM-Math~\cite{ying2024internlmmath}. Future work could explore adaptive hierarchies and the cross-domain generalization.
\section*{Ethical Considerations}
Our work focuses on improving theorem proving through Hierarchical Attention while addressing several ethical considerations. We use publicly available datasets, including LeanDojo Benchmark 4 under the MIT license\footnote{\url{https://github.com/lean-dojo/LeanDojo/blob/main/LICENSE}}, and strictly follow data usage policies. While mathematical content is generally objective, we acknowledge potential biases in theorem selection and proof styles. Our method, though designed for positive applications, should be used with human oversight as it could potentially generate misleading proofs. To promote transparency and reproducibility, we will release our code and models with appropriate licenses and usage guidelines.

\bibliography{acl_latex}

\begin{thebibliography}{76}
\providecommand{\natexlab}[1]{#1}

\bibitem[{Abdelaziz et~al.(2022)Abdelaziz, Crouse, Makni, Austel, Cornelio, Ikbal, Kapanipathi, Makondo, Srinivas, Witbrock et~al.}]{abdelaziz2022learning}
Ibrahim Abdelaziz, Maxwell Crouse, Bassem Makni, Vernon Austel, Cristina Cornelio, Shajith Ikbal, Pavan Kapanipathi, Ndivhuwo Makondo, Kavitha Srinivas, Michael Witbrock, et~al. 2022.
\newblock Learning to guide a saturation-based theorem prover.
\newblock \emph{IEEE Transactions on Pattern Analysis and Machine Intelligence}, 45(1):738--751.

\bibitem[{Ayg{\"u}n et~al.(2020)Ayg{\"u}n, Ahmed, Anand, Firoiu, Glorot, Orseau, Precup, and Mourad}]{aygun2020learning}
Eser Ayg{\"u}n, Zafarali Ahmed, Ankit Anand, Vlad Firoiu, Xavier Glorot, Laurent Orseau, Doina Precup, and Shibl Mourad. 2020.
\newblock Learning to prove from synthetic theorems.
\newblock \emph{arXiv preprint arXiv:2006.11259}.

\bibitem[{Ayg{\"u}n et~al.(2022)Ayg{\"u}n, Anand, Orseau, Glorot, Mcaleer, Firoiu, Zhang, Precup, and Mourad}]{aygun2022proving}
Eser Ayg{\"u}n, Ankit Anand, Laurent Orseau, Xavier Glorot, Stephen~M Mcaleer, Vlad Firoiu, Lei~M Zhang, Doina Precup, and Shibl Mourad. 2022.
\newblock Proving theorems using incremental learning and hindsight experience replay.
\newblock In \emph{International Conference on Machine Learning}, pages 1198--1210. PMLR.

\bibitem[{Azerbayev et~al.(2023)Azerbayev, Piotrowski, Schoelkopf, Ayers, Radev, and Avigad}]{azerbayev2023proofnet}
Zhangir Azerbayev, Bartosz Piotrowski, Hailey Schoelkopf, Edward~W Ayers, Dragomir Radev, and Jeremy Avigad. 2023.
\newblock Proofnet: Autoformalizing and formally proving undergraduate-level mathematics.
\newblock \emph{arXiv preprint arXiv:2302.12433}.

\bibitem[{Bauer et~al.(2024)Bauer, Petkovi{\'c}, and Todorovski}]{bauer2024mlfmf}
Andrej Bauer, Matej Petkovi{\'c}, and Ljupco Todorovski. 2024.
\newblock Mlfmf: data sets for machine learning for mathematical formalization.
\newblock \emph{Advances in Neural Information Processing Systems}, 36.

\bibitem[{Biderman et~al.(2023)Biderman, Schoelkopf, Anthony, Bradley, O’Brien, Hallahan, Khan, Purohit, Prashanth, Raff et~al.}]{biderman2023pythia}
Stella Biderman, Hailey Schoelkopf, Quentin~Gregory Anthony, Herbie Bradley, Kyle O’Brien, Eric Hallahan, Mohammad~Aflah Khan, Shivanshu Purohit, USVSN~Sai Prashanth, Edward Raff, et~al. 2023.
\newblock Pythia: A suite for analyzing large language models across training and scaling.
\newblock In \emph{International Conference on Machine Learning}, pages 2397--2430. PMLR.

\bibitem[{Chou et~al.(2000)Chou, Gao, and Zhang}]{chou2000deductive}
Shang-Ching Chou, Xiao-Shan Gao, and Jing-Zhong Zhang. 2000.
\newblock A deductive database approach to automated geometry theorem proving and discovering.
\newblock \emph{Journal of Automated Reasoning}, 25(3):219--246.

\bibitem[{Chvalovsk{\`y} et~al.(2023)Chvalovsk{\`y}, Korovin, Piepenbrock, and Urban}]{chvalovsky2023guiding}
Karel Chvalovsk{\`y}, Konstantin Korovin, Jelle Piepenbrock, and Josef Urban. 2023.
\newblock Guiding an instantiation prover with graph neural networks.
\newblock In \emph{LPAR}, pages 112--123.

\bibitem[{Crouse et~al.(2021)Crouse, Abdelaziz, Makni, Whitehead, Cornelio, Kapanipathi, Srinivas, Thost, Witbrock, and Fokoue}]{crouse2021deep}
Maxwell Crouse, Ibrahim Abdelaziz, Bassem Makni, Spencer Whitehead, Cristina Cornelio, Pavan Kapanipathi, Kavitha Srinivas, Veronika Thost, Michael Witbrock, and Achille Fokoue. 2021.
\newblock A deep reinforcement learning approach to first-order logic theorem proving.
\newblock In \emph{Proceedings of the AAAI Conference on Artificial Intelligence}, volume~35, pages 6279--6287.

\bibitem[{De~Moura and Bj{\o}rner(2008)}]{de2008z3}
Leonardo De~Moura and Nikolaj Bj{\o}rner. 2008.
\newblock Z3: An efficient smt solver.
\newblock In \emph{International conference on Tools and Algorithms for the Construction and Analysis of Systems}, pages 337--340. Springer.

\bibitem[{De~Moura et~al.(2015)De~Moura, Kong, Avigad, Van~Doorn, and von Raumer}]{de2015lean}
Leonardo De~Moura, Soonho Kong, Jeremy Avigad, Floris Van~Doorn, and Jakob von Raumer. 2015.
\newblock The lean theorem prover (system description).
\newblock In \emph{Automated Deduction-CADE-25: 25th International Conference on Automated Deduction, Berlin, Germany, August 1-7, 2015, Proceedings 25}, pages 378--388. Springer.

\bibitem[{Dong and Ma(2025)}]{dong2025stpselfplayllmtheorem}
Kefan Dong and Tengyu Ma. 2025.
\newblock \href {https://arxiv.org/abs/2502.00212} {Stp: Self-play llm theorem provers with iterative conjecturing and proving}.
\newblock \emph{Preprint}, arXiv:2502.00212.

\bibitem[{E{\'e}n and S{\"o}rensson(2003)}]{een2003extensible}
Niklas E{\'e}n and Niklas S{\"o}rensson. 2003.
\newblock An extensible sat-solver.
\newblock In \emph{International conference on theory and applications of satisfiability testing}, pages 502--518. Springer.

\bibitem[{Ferreira and Freitas(2020{\natexlab{a}})}]{ferreira2020natural}
Deborah Ferreira and Andr{\'e} Freitas. 2020{\natexlab{a}}.
\newblock Natural language premise selection: Finding supporting statements for mathematical text.
\newblock \emph{arXiv preprint arXiv:2004.14959}.

\bibitem[{Ferreira and Freitas(2020{\natexlab{b}})}]{ferreira2020premise}
Deborah Ferreira and Andr{\'e} Freitas. 2020{\natexlab{b}}.
\newblock Premise selection in natural language mathematical texts.
\newblock In \emph{Proceedings of the 58th Annual Meeting of the Association for Computational Linguistics}, pages 7365--7374.

\bibitem[{First et~al.(2023)First, Rabe, Ringer, and Brun}]{first2023baldur}
Emily First, Markus~N Rabe, Talia Ringer, and Yuriy Brun. 2023.
\newblock Baldur: Whole-proof generation and repair with large language models.
\newblock In \emph{Proceedings of the 31st ACM Joint European Software Engineering Conference and Symposium on the Foundations of Software Engineering}, pages 1229--1241.

\bibitem[{Fokoue et~al.(2023)Fokoue, Abdelaziz, Crouse, Ikbal, Kishimoto, Lima, Makondo, and Marinescu}]{fokoue2023ensemble}
Achille Fokoue, Ibrahim Abdelaziz, Maxwell Crouse, Shajith Ikbal, Akihiro Kishimoto, Guilherme Lima, Ndivhuwo Makondo, and Radu Marinescu. 2023.
\newblock An ensemble approach for automated theorem proving based on efficient name invariant graph neural representations.
\newblock \emph{arXiv preprint arXiv:2305.08676}.

\bibitem[{Han et~al.(2021)Han, Rute, Wu, Ayers, and Polu}]{han2021proof}
Jesse~Michael Han, Jason Rute, Yuhuai Wu, Edward~W Ayers, and Stanislas Polu. 2021.
\newblock Proof artifact co-training for theorem proving with language models.
\newblock \emph{arXiv preprint arXiv:2102.06203}.

\bibitem[{Holden and Korovin(2025)}]{holden2025graph}
Edvard~K Holden and Konstantin Korovin. 2025.
\newblock Graph sequence learning for premise selection.
\newblock \emph{Journal of Symbolic Computation}, 128:102376.

\bibitem[{Huang et~al.(2018)Huang, Dhariwal, Song, and Sutskever}]{huang2018gamepad}
Daniel Huang, Prafulla Dhariwal, Dawn Song, and Ilya Sutskever. 2018.
\newblock Gamepad: A learning environment for theorem proving.
\newblock \emph{arXiv preprint arXiv:1806.00608}.

\bibitem[{Irving et~al.(2016)Irving, Szegedy, Alemi, E{\'e}n, Chollet, and Urban}]{irving2016deepmath}
Geoffrey Irving, Christian Szegedy, Alexander~A Alemi, Niklas E{\'e}n, Fran{\c{c}}ois Chollet, and Josef Urban. 2016.
\newblock Deepmath-deep sequence models for premise selection.
\newblock \emph{Advances in neural information processing systems}, 29.

\bibitem[{Jiang et~al.(2023)Jiang, Li, and Jamnik}]{jiang2023multilingual}
Albert~Q Jiang, Wenda Li, and Mateja Jamnik. 2023.
\newblock Multilingual mathematical autoformalization.
\newblock \emph{arXiv preprint arXiv:2311.03755}.

\bibitem[{Jiang et~al.(2022{\natexlab{a}})Jiang, Welleck, Zhou, Li, Liu, Jamnik, Lacroix, Wu, and Lample}]{jiang2022draft}
Albert~Q Jiang, Sean Welleck, Jin~Peng Zhou, Wenda Li, Jiacheng Liu, Mateja Jamnik, Timoth{\'e}e Lacroix, Yuhuai Wu, and Guillaume Lample. 2022{\natexlab{a}}.
\newblock Draft, sketch, and prove: Guiding formal theorem provers with informal proofs.
\newblock \emph{arXiv preprint arXiv:2210.12283}.

\bibitem[{Jiang et~al.(2021)Jiang, Li, Han, and Wu}]{jiang2021lisa}
Albert~Qiaochu Jiang, Wenda Li, Jesse~Michael Han, and Yuhuai Wu. 2021.
\newblock Lisa: Language models of isabelle proofs.
\newblock In \emph{6th Conference on Artificial Intelligence and Theorem Proving}, pages 378--392.

\bibitem[{Jiang et~al.(2022{\natexlab{b}})Jiang, Li, Tworkowski, Czechowski, Odrzyg{\'o}{\'z}d{\'z}, Mi{\l}o{\'s}, Wu, and Jamnik}]{jiang2022thor}
Albert~Qiaochu Jiang, Wenda Li, Szymon Tworkowski, Konrad Czechowski, Tomasz Odrzyg{\'o}{\'z}d{\'z}, Piotr Mi{\l}o{\'s}, Yuhuai Wu, and Mateja Jamnik. 2022{\natexlab{b}}.
\newblock Thor: Wielding hammers to integrate language models and automated theorem provers.
\newblock \emph{Advances in Neural Information Processing Systems}, 35:8360--8373.

\bibitem[{Kov{\'a}cs and Voronkov(2013)}]{kovacs2013first}
Laura Kov{\'a}cs and Andrei Voronkov. 2013.
\newblock First-order theorem proving and vampire.
\newblock In \emph{International Conference on Computer Aided Verification}, pages 1--35. Springer.

\bibitem[{Kucik and Korovin(2018)}]{kucik2018premise}
Andrzej~Stanis{\l}aw Kucik and Konstantin Korovin. 2018.
\newblock Premise selection with neural networks and distributed representation of features.
\newblock \emph{arXiv preprint arXiv:1807.10268}.

\bibitem[{Lample et~al.(2022)Lample, Lacroix, Lachaux, Rodriguez, Hayat, Lavril, Ebner, and Martinet}]{lample2022hypertree}
Guillaume Lample, Timothee Lacroix, Marie-Anne Lachaux, Aurelien Rodriguez, Amaury Hayat, Thibaut Lavril, Gabriel Ebner, and Xavier Martinet. 2022.
\newblock Hypertree proof search for neural theorem proving.
\newblock \emph{Advances in neural information processing systems}, 35:26337--26349.

\bibitem[{Lin et~al.(2024)Lin, Sun, Yang, and Welleck}]{lin2024lean}
Haohan Lin, Zhiqing Sun, Yiming Yang, and Sean Welleck. 2024.
\newblock Lean-star: Learning to interleave thinking and proving.
\newblock \emph{arXiv preprint arXiv:2407.10040}.

\bibitem[{Lin et~al.(2025)Lin, Tang, Lyu, Wu, Lin, Yang, Li, Xia, Chen, Arora, and Jin}]{lin2025goedelproverfrontiermodelopensource}
Yong Lin, Shange Tang, Bohan Lyu, Jiayun Wu, Hongzhou Lin, Kaiyu Yang, Jia Li, Mengzhou Xia, Danqi Chen, Sanjeev Arora, and Chi Jin. 2025.
\newblock \href {https://arxiv.org/abs/2502.07640} {Goedel-prover: A frontier model for open-source automated theorem proving}.
\newblock \emph{Preprint}, arXiv:2502.07640.

\bibitem[{Liu et~al.(2023)Liu, Shen, Xin, Liu, Yuan, Wang, Ju, Zheng, Yin, Li et~al.}]{liu2023fimo}
Chengwu Liu, Jianhao Shen, Huajian Xin, Zhengying Liu, Ye~Yuan, Haiming Wang, Wei Ju, Chuanyang Zheng, Yichun Yin, Lin Li, et~al. 2023.
\newblock Fimo: A challenge formal dataset for automated theorem proving.
\newblock \emph{arXiv preprint arXiv:2309.04295}.

\bibitem[{Loos et~al.(2017)Loos, Irving, Szegedy, and Kaliszyk}]{loos2017deep}
Sarah Loos, Geoffrey Irving, Christian Szegedy, and Cezary Kaliszyk. 2017.
\newblock Deep network guided proof search.
\newblock \emph{arXiv preprint arXiv:1701.06972}.

\bibitem[{Lu et~al.(2024)Lu, Wan, Liu, Huang, Xiong, Liu, Shen, Jin, Zhang, Wang et~al.}]{lu2024process}
Jianqiao Lu, Yingjia Wan, Zhengying Liu, Yinya Huang, Jing Xiong, Chengwu Liu, Jianhao Shen, Hui Jin, Jipeng Zhang, Haiming Wang, et~al. 2024.
\newblock Process-driven autoformalization in lean 4.
\newblock \emph{arXiv preprint arXiv:2406.01940}.

\bibitem[{McKeown and Sutcliffe(2023)}]{mckeown2023reinforcement}
Jack McKeown and Geoff Sutcliffe. 2023.
\newblock Reinforcement learning for guiding the e theorem prover.
\newblock In \emph{The International FLAIRS Conference Proceedings}, volume~36.

\bibitem[{Miku{\l}a et~al.(2023)Miku{\l}a, Tworkowski, Antoniak, Piotrowski, Jiang, Zhou, Szegedy, Kuci{\'n}ski, Mi{\l}o{\'s}, and Wu}]{mikula2023magnushammer}
Maciej Miku{\l}a, Szymon Tworkowski, Szymon Antoniak, Bartosz Piotrowski, Albert~Qiaochu Jiang, Jin~Peng Zhou, Christian Szegedy, {\L}ukasz Kuci{\'n}ski, Piotr Mi{\l}o{\'s}, and Yuhuai Wu. 2023.
\newblock Magnushammer: A transformer-based approach to premise selection.
\newblock \emph{arXiv preprint arXiv:2303.04488}.

\bibitem[{Moura and Ullrich(2021)}]{moura2021lean}
Leonardo~de Moura and Sebastian Ullrich. 2021.
\newblock The lean 4 theorem prover and programming language.
\newblock In \emph{Automated Deduction--CADE 28: 28th International Conference on Automated Deduction, Virtual Event, July 12--15, 2021, Proceedings 28}, pages 625--635. Springer.

\bibitem[{Murphy et~al.(2024)Murphy, Yang, Sun, Li, Anandkumar, and Si}]{murphy2024autoformalizing}
Logan Murphy, Kaiyu Yang, Jialiang Sun, Zhaoyu Li, Anima Anandkumar, and Xujie Si. 2024.
\newblock Autoformalizing euclidean geometry.
\newblock \emph{arXiv preprint arXiv:2405.17216}.

\bibitem[{Paliwal et~al.(2020)Paliwal, Loos, Rabe, Bansal, and Szegedy}]{paliwal2020graph}
Aditya Paliwal, Sarah Loos, Markus Rabe, Kshitij Bansal, and Christian Szegedy. 2020.
\newblock Graph representations for higher-order logic and theorem proving.
\newblock In \emph{Proceedings of the AAAI Conference on Artificial Intelligence}, volume~34, pages 2967--2974.

\bibitem[{Paulson(1994)}]{paulson1994isabelle}
Lawrence~C Paulson. 1994.
\newblock \emph{Isabelle: A generic theorem prover}.
\newblock Springer.

\bibitem[{Peng and Ma(2017)}]{peng2017tree}
Kebin Peng and Dianfu Ma. 2017.
\newblock Tree-structure cnn for automated theorem proving.
\newblock In \emph{Neural Information Processing: 24th International Conference, ICONIP 2017, Guangzhou, China, November 14-18, 2017, Proceedings, Part II 24}, pages 3--12. Springer.

\bibitem[{Piotrowski and Urban(2020)}]{piotrowski2020stateful}
Bartosz Piotrowski and Josef Urban. 2020.
\newblock Stateful premise selection by recurrent neural networks.
\newblock \emph{arXiv preprint arXiv:2004.08212}.

\bibitem[{Polu et~al.(2022)Polu, Han, Zheng, Baksys, Babuschkin, and Sutskever}]{polu2022formal}
Stanislas Polu, Jesse~Michael Han, Kunhao Zheng, Mantas Baksys, Igor Babuschkin, and Ilya Sutskever. 2022.
\newblock Formal mathematics statement curriculum learning.
\newblock \emph{arXiv preprint arXiv:2202.01344}.

\bibitem[{Polu and Sutskever(2020)}]{polu2020generative}
Stanislas Polu and Ilya Sutskever. 2020.
\newblock Generative language modeling for automated theorem proving.
\newblock \emph{arXiv preprint arXiv:2009.03393}.

\bibitem[{Rawson and Reger(2019)}]{rawson2019neurally}
Michael Rawson and Giles Reger. 2019.
\newblock A neurally-guided, parallel theorem prover.
\newblock In \emph{Frontiers of Combining Systems: 12th International Symposium, FroCoS 2019, London, UK, September 4-6, 2019, Proceedings 12}, pages 40--56. Springer.

\bibitem[{Rawson and Reger(2020)}]{rawson2020directed}
Michael Rawson and Giles Reger. 2020.
\newblock Directed graph networks for logical reasoning.
\newblock In \emph{PAAR+ SC$^2$@ IJCAR}, pages 109--119.

\bibitem[{Rawson and Reger(2021)}]{rawson2021lazycop}
Michael Rawson and Giles Reger. 2021.
\newblock lazycop: Lazy paramodulation meets neurally guided search.
\newblock In \emph{Automated Reasoning with Analytic Tableaux and Related Methods: 30th International Conference, TABLEAUX 2021, Birmingham, UK, September 6--9, 2021, Proceedings 30}, pages 187--199. Springer.

\bibitem[{Rute et~al.(2024)Rute, Ol{\v{s}}{\'a}k, Blaauwbroek, Massolo, Piepenbrock, and Pestun}]{rute2024graph2tac}
Jason Rute, Miroslav Ol{\v{s}}{\'a}k, Lasse Blaauwbroek, Fidel Ivan~Schaposnik Massolo, Jelle Piepenbrock, and Vasily Pestun. 2024.
\newblock Graph2tac: learning hierarchical representations of math concepts in theorem proving.
\newblock \emph{arXiv preprint arXiv:2401.02949}.

\bibitem[{Sanchez-Stern et~al.(2020)Sanchez-Stern, Alhessi, Saul, and Lerner}]{sanchez2020generating}
Alex Sanchez-Stern, Yousef Alhessi, Lawrence Saul, and Sorin Lerner. 2020.
\newblock Generating correctness proofs with neural networks.
\newblock In \emph{Proceedings of the 4th ACM SIGPLAN International Workshop on Machine Learning and Programming Languages}, pages 1--10.

\bibitem[{Sanchez-Stern et~al.(2023)Sanchez-Stern, First, Zhou, Kaufman, Brun, and Ringer}]{sanchez2023passport}
Alex Sanchez-Stern, Emily First, Timothy Zhou, Zhanna Kaufman, Yuriy Brun, and Talia Ringer. 2023.
\newblock Passport: Improving automated formal verification using identifiers.
\newblock \emph{ACM Transactions on Programming Languages and Systems}, 45(2):1--30.

\bibitem[{Schulz(2002)}]{schulz2002brainiac}
Stephan Schulz. 2002.
\newblock E--a brainiac theorem prover.
\newblock \emph{Ai Communications}, 15(2-3):111--126.

\bibitem[{Suda(2021)}]{suda2021improving}
Martin Suda. 2021.
\newblock Improving enigma-style clause selection while learning from history.
\newblock In \emph{Automated Deduction--CADE 28: 28th International Conference on Automated Deduction, Virtual Event, July 12--15, 2021, Proceedings 28}, pages 543--561. Springer.

\bibitem[{{The Coq Development Team}(2024)}]{coq}
{The Coq Development Team}. 2024.
\newblock \emph{Coq}.
\newblock URL \url{https://coq.inria.fr}.

\bibitem[{Vaswani(2017)}]{vaswani2017attention}
A~Vaswani. 2017.
\newblock Attention is all you need.
\newblock \emph{Advances in Neural Information Processing Systems}.

\bibitem[{Wang et~al.(2025{\natexlab{a}})Wang, Unsal, Lin, Baksys, Liu, Santos, Sung, Vinyes, Ying, Zhu, Lu, de~Saxcé, Bailey, Song, Xiao, Zhang, Zhang, Pu, Zhu, Liu, Bayer, Michel, Yu, Dreyfus-Schmidt, Tunstall, Pagani, Machado, Bourigault, Wang, Polu, Barroyer, Li, Niu, Fleureau, Hu, Yu, Wang, Yang, Liu, and Li}]{wang2025kiminaproverpreviewlargeformal}
Haiming Wang, Mert Unsal, Xiaohan Lin, Mantas Baksys, Junqi Liu, Marco~Dos Santos, Flood Sung, Marina Vinyes, Zhenzhe Ying, Zekai Zhu, Jianqiao Lu, Hugues de~Saxcé, Bolton Bailey, Chendong Song, Chenjun Xiao, Dehao Zhang, Ebony Zhang, Frederick Pu, Han Zhu, Jiawei Liu, Jonas Bayer, Julien Michel, Longhui Yu, Léo Dreyfus-Schmidt, Lewis Tunstall, Luigi Pagani, Moreira Machado, Pauline Bourigault, Ran Wang, Stanislas Polu, Thibaut Barroyer, Wen-Ding Li, Yazhe Niu, Yann Fleureau, Yangyang Hu, Zhouliang Yu, Zihan Wang, Zhilin Yang, Zhengying Liu, and Jia Li. 2025{\natexlab{a}}.
\newblock \href {https://arxiv.org/abs/2504.11354} {Kimina-prover preview: Towards large formal reasoning models with reinforcement learning}.
\newblock \emph{Preprint}, arXiv:2504.11354.

\bibitem[{Wang et~al.(2024)Wang, Xin, Liu, Li, Huang, Lu, Yang, Tang, Yin, Li et~al.}]{wang2024proving}
Haiming Wang, Huajian Xin, Zhengying Liu, Wenda Li, Yinya Huang, Jianqiao Lu, Zhicheng Yang, Jing Tang, Jian Yin, Zhenguo Li, et~al. 2024.
\newblock Proving theorems recursively.
\newblock \emph{arXiv preprint arXiv:2405.14414}.

\bibitem[{Wang et~al.(2023{\natexlab{a}})Wang, Xin, Zheng, Li, Liu, Cao, Huang, Xiong, Shi, Xie et~al.}]{wang2023lego}
Haiming Wang, Huajian Xin, Chuanyang Zheng, Lin Li, Zhengying Liu, Qingxing Cao, Yinya Huang, Jing Xiong, Han Shi, Enze Xie, et~al. 2023{\natexlab{a}}.
\newblock Lego-prover: Neural theorem proving with growing libraries.
\newblock \emph{arXiv preprint arXiv:2310.00656}.

\bibitem[{Wang et~al.(2023{\natexlab{b}})Wang, Yuan, Liu, Shen, Yin, Xiong, Xie, Shi, Li, Li et~al.}]{wang2023dt}
Haiming Wang, Ye~Yuan, Zhengying Liu, Jianhao Shen, Yichun Yin, Jing Xiong, Enze Xie, Han Shi, Yujun Li, Lin Li, et~al. 2023{\natexlab{b}}.
\newblock Dt-solver: Automated theorem proving with dynamic-tree sampling guided by proof-level value function.
\newblock In \emph{Proceedings of the 61st Annual Meeting of the Association for Computational Linguistics (Volume 1: Long Papers)}, pages 12632--12646.

\bibitem[{Wang et~al.(2017)Wang, Tang, Wang, and Deng}]{wang2017premise}
Mingzhe Wang, Yihe Tang, Jian Wang, and Jia Deng. 2017.
\newblock Premise selection for theorem proving by deep graph embedding.
\newblock \emph{Advances in neural information processing systems}, 30.

\bibitem[{Wang et~al.(2020)Wang, Brown, Kaliszyk, and Urban}]{wang2020exploration}
Qingxiang Wang, Chad Brown, Cezary Kaliszyk, and Josef Urban. 2020.
\newblock Exploration of neural machine translation in autoformalization of mathematics in mizar.
\newblock In \emph{Proceedings of the 9th ACM SIGPLAN International Conference on Certified Programs and Proofs}, pages 85--98.

\bibitem[{Wang et~al.(2018)Wang, Kaliszyk, and Urban}]{wang2018first}
Qingxiang Wang, Cezary Kaliszyk, and Josef Urban. 2018.
\newblock First experiments with neural translation of informal to formal mathematics.
\newblock In \emph{Intelligent Computer Mathematics: 11th International Conference, CICM 2018, Hagenberg, Austria, August 13-17, 2018, Proceedings 11}, pages 255--270. Springer.

\bibitem[{Wang et~al.(2025{\natexlab{b}})Wang, Pan, Li, Zhang, Jia, Diao, Pi, Hu, and Zhang}]{wang2025malotmultiagentleanbasedlong}
Ruida Wang, Rui Pan, Yuxin Li, Jipeng Zhang, Yizhen Jia, Shizhe Diao, Renjie Pi, Junjie Hu, and Tong Zhang. 2025{\natexlab{b}}.
\newblock \href {https://arxiv.org/abs/2503.03205} {Ma-lot: Multi-agent lean-based long chain-of-thought reasoning enhances formal theorem proving}.
\newblock \emph{Preprint}, arXiv:2503.03205.

\bibitem[{Welleck and Saha(2023)}]{welleck2023llmstep}
Sean Welleck and Rahul Saha. 2023.
\newblock Llmstep: Llm proofstep suggestions in lean.
\newblock \emph{arXiv preprint arXiv:2310.18457}.

\bibitem[{Wu et~al.(2022{\natexlab{a}})Wu, Zhang, and Huang}]{wu2022automatic}
Qinzhuo Wu, Qi~Zhang, and Xuanjing Huang. 2022{\natexlab{a}}.
\newblock Automatic math word problem generation with topic-expression co-attention mechanism and reinforcement learning.
\newblock \emph{IEEE/ACM Transactions on Audio, Speech, and Language Processing}, 30:1061--1072.

\bibitem[{Wu(2022)}]{wu2022formal}
Yuhuai Wu. 2022.
\newblock Formal premise selection with language models.
\newblock In \emph{Conference on Artificial Intelligence and Theorem Proving (AITP)}, volume~4.

\bibitem[{Wu et~al.(2022{\natexlab{b}})Wu, Jiang, Li, Rabe, Staats, Jamnik, and Szegedy}]{wu2022autoformalization}
Yuhuai Wu, Albert~Qiaochu Jiang, Wenda Li, Markus Rabe, Charles Staats, Mateja Jamnik, and Christian Szegedy. 2022{\natexlab{b}}.
\newblock Autoformalization with large language models.
\newblock \emph{Advances in Neural Information Processing Systems}, 35:32353--32368.

\bibitem[{Wu et~al.(2024)Wu, Wang, Lin, and Chen}]{wu2024lean}
Zijian Wu, Jiayu Wang, Dahua Lin, and Kai Chen. 2024.
\newblock Lean-github: Compiling github lean repositories for a versatile lean prover.
\newblock \emph{arXiv preprint arXiv:2407.17227}.

\bibitem[{Xin et~al.(2024)Xin, Guo, Shao, Ren, Zhu, Liu, Ruan, Li, and Liang}]{xin2024deepseek}
Huajian Xin, Daya Guo, Zhihong Shao, Zhizhou Ren, Qihao Zhu, Bo~Liu, Chong Ruan, Wenda Li, and Xiaodan Liang. 2024.
\newblock Deepseek-prover: Advancing theorem proving in llms through large-scale synthetic data.
\newblock \emph{arXiv preprint arXiv:2405.14333}.

\bibitem[{Xin et~al.(2025)Xin, Xi, Yang, Chen, Wu, Xiao, Sun, Zheng, and Shen}]{xin2025bfsproverscalablebestfirsttree}
Ran Xin, Chenguang Xi, Jie Yang, Feng Chen, Hang Wu, Xia Xiao, Yifan Sun, Shen Zheng, and Kai Shen. 2025.
\newblock \href {https://arxiv.org/abs/2502.03438} {Bfs-prover: Scalable best-first tree search for llm-based automatic theorem proving}.
\newblock \emph{Preprint}, arXiv:2502.03438.

\bibitem[{Yang and Deng(2019)}]{yang2019learning}
Kaiyu Yang and Jia Deng. 2019.
\newblock Learning to prove theorems via interacting with proof assistants.
\newblock In \emph{International Conference on Machine Learning}, pages 6984--6994. PMLR.

\bibitem[{Yang et~al.(2024)Yang, Swope, Gu, Chalamala, Song, Yu, Godil, Prenger, and Anandkumar}]{yang2024leandojo}
Kaiyu Yang, Aidan Swope, Alex Gu, Rahul Chalamala, Peiyang Song, Shixing Yu, Saad Godil, Ryan~J Prenger, and Animashree Anandkumar. 2024.
\newblock Leandojo: Theorem proving with retrieval-augmented language models.
\newblock \emph{Advances in Neural Information Processing Systems}, 36.

\bibitem[{Yang et~al.(2016)Yang, Yang, Dyer, He, Smola, and Hovy}]{yang2016hierarchical}
Zichao Yang, Diyi Yang, Chris Dyer, Xiaodong He, Alex Smola, and Eduard Hovy. 2016.
\newblock Hierarchical attention networks for document classification.
\newblock In \emph{Proceedings of the 2016 conference of the North American chapter of the association for computational linguistics: human language technologies}, pages 1480--1489.

\bibitem[{Ying et~al.(2024{\natexlab{a}})Ying, Wu, Geng, Wang, Lin, and Chen}]{ying2024lean}
Huaiyuan Ying, Zijian Wu, Yihan Geng, Jiayu Wang, Dahua Lin, and Kai Chen. 2024{\natexlab{a}}.
\newblock Lean workbook: A large-scale lean problem set formalized from natural language math problems.
\newblock \emph{arXiv preprint arXiv:2406.03847}.

\bibitem[{Ying et~al.(2024{\natexlab{b}})Ying, Zhang, Li, Zhou, Shao, Fei, Ma, Hong, Liu, Wang, Wang, Wu, Li, Zhou, Liu, Zhang, Zhang, Yan, Qiu, Wang, Chen, and Lin}]{ying2024internlmmath}
Huaiyuan Ying, Shuo Zhang, Linyang Li, Zhejian Zhou, Yunfan Shao, Zhaoye Fei, Yichuan Ma, Jiawei Hong, Kuikun Liu, Ziyi Wang, Yudong Wang, Zijian Wu, Shuaibin Li, Fengzhe Zhou, Hongwei Liu, Songyang Zhang, Wenwei Zhang, Hang Yan, Xipeng Qiu, Jiayu Wang, Kai Chen, and Dahua Lin. 2024{\natexlab{b}}.
\newblock \href {https://arxiv.org/abs/2402.06332} {Internlm-math: Open math large language models toward verifiable reasoning}.
\newblock \emph{Preprint}, arXiv:2402.06332.

\bibitem[{Zhang et~al.(2025)Zhang, Wang, Ji, Liu, Yue, Zhang, Zhang, Zhou, and Gai}]{zhang2025leanabellproverposttrainingscalingformal}
Jingyuan Zhang, Qi~Wang, Xingguang Ji, Yahui Liu, Yang Yue, Fuzheng Zhang, Di~Zhang, Guorui Zhou, and Kun Gai. 2025.
\newblock \href {https://arxiv.org/abs/2504.06122} {Leanabell-prover: Posttraining scaling in formal reasoning}.
\newblock \emph{Preprint}, arXiv:2504.06122.

\bibitem[{Zhao et~al.(2023)Zhao, Li, and Kong}]{zhao2023decomposing}
Xueliang Zhao, Wenda Li, and Lingpeng Kong. 2023.
\newblock Decomposing the enigma: Subgoal-based demonstration learning for formal theorem proving.
\newblock \emph{arXiv preprint arXiv:2305.16366}.

\bibitem[{Zheng et~al.(2021)Zheng, Han, and Polu}]{zheng2021minif2f}
Kunhao Zheng, Jesse~Michael Han, and Stanislas Polu. 2021.
\newblock Minif2f: a cross-system benchmark for formal olympiad-level mathematics.
\newblock \emph{arXiv preprint arXiv:2109.00110}.

\end{thebibliography}

\clearpage
\appendix
\section{Appendix}

\subsection{Training Details}
\label{app:training_details}

We use Pythia-2.8B\footnote{\url{https://huggingface.co/EleutherAI/pythia-2.8b}} as our base model. The training data is sourced from LeanDojo Benchmark 4 \footnote{Yang, K. (2023). LeanDojo Benchmark (v1) [Data set]. Zenodo. \url{https://doi.org/10.5281/zenodo.8016386}}, which consists of 169,530 samples for training and 3,606 samples for validation.

We train the model for 3 epochs on 8 NVIDIA A800 GPUs using DeepSpeed\footnote{\url{https://github.com/microsoft/DeepSpeed}} with ZeRO-3 optimization, taking approximately 40 hours. The training uses a per-device batch size of 2 with gradient accumulation steps of 2, resulting in an effective batch size of 32. We adopt a learning rate of 1e-5 with a cosine decay schedule and 3\% warmup ratio. The training process employs FP16 precision without weight decay, and ZeRO-3 is configured with parameter and optimizer state partitioning across GPUs. For reproducibility, we set the random seed to 42 across all experiments.

During training, we evaluate the model every 500 steps and save checkpoints at the same frequency, maintaining the 3 most recent checkpoints. The best model is selected based on validation performance at the end of training. 
The training objective combines the standard cross-entropy loss with our hierarchical flow loss. Table~\ref{tab:hyperparams} shows the specific hyperparameters ($\lambda$ and $L$) used for different evaluation sets.

\begin{table}[h]
\begin{center}
\caption{Hyperparameters for different evaluation sets.}
\label{tab:hyperparams}
\scalebox{0.75}{
\begin{tabular}{lcc}
\toprule
Dataset & $\lambda$ & $L$ \\
\midrule
miniF2F (test) & 0.1 & 4 \\
miniF2F (valid) & 0.1 & 16 \\
ProofNet (test) & 0.2 & 16 \\
ProofNet (valid) & 0.2 & 4 \\
\bottomrule
\end{tabular}
}
\end{center}
\end{table}

\subsection{Evaluation algorithm}
\label{app:evaluation_algorithm}

We implement two evaluation algorithms for theorem proving: best-first search and single-pass sampling. Both algorithms share the same computation budget $K \times T$, where $T=100$ is the maximum expansion steps.

\paragraph{Best-First Search} maintains a priority queue of states ranked by trajectory score $\sum_{j=0}^{i-1} \log p(a_j|s_j)$. For each expansion, it selects the highest-scoring state $s_i$, generates $S$ candidate tactics, and creates new states by applying valid tactics. The search succeeds when reaching a state with no remaining goals within $N$ expansions.

\paragraph{Single-Pass Sampling} runs $K$ parallel proof attempts. Each attempt samples tactics sequentially until finding a valid one or reaching the attempt limit. A proof succeeds if it completes within $N$ valid tactics. This approach simplifies the search process by setting $S=1$ and focusing on trajectory sampling rather than state ranking.

\subsection{Ablation Studies}
\subsubsection{Layer-wise Adaptation Mechanism}
To validate the effectiveness of our layer-wise adaptation mechanism ($\alpha_l = 1-l/L$), we conduct ablation studies on miniF2F and ProofNet benchmarks using best-first search with $K = 64$. The results are shown in Table~\ref{tab:ablation}.

\begin{table}[h!]
\begin{center}
\caption{Ablation study results on miniF2F and ProofNet benchmarks with best-first search ($K = 64$).}
\label{tab:ablation}
\scalebox{0.75}{
\begin{tabular}{lcccc}
\toprule
\multirow{2}{*}{Method} & \multicolumn{2}{c}{miniF2F} & \multicolumn{2}{c}{ProofNet} \\
\cmidrule(lr){2-3} \cmidrule(lr){4-5}
& Test & Valid & Test & Valid \\
\midrule
PASS (baseline) & 29.51 & 31.56 & 13.56 & 10.17 \\
PASS (w/o adaptation) & 30.74 & 32.34 & 14.69 & 11.30 \\
PASS (w/ adaptation) & \textbf{31.56} & \textbf{34.02} & \textbf{15.25} & \textbf{11.86} \\
\midrule
$R_{avg}$ (w/o adaptation) & \textbf{0.53} & 0.82 & \textbf{0.69} & \textbf{0.50} \\
$R_{avg}$ (w/ adaptation) & 0.76 & \textbf{0.64} & 0.84 & \textbf{0.50} \\
$\text{Diff.}$(w/o adaptation) (\%) & 9.86 & 11.27 & 22.73 & 18.75 \\
$\text{Diff.}$(w/ adaptation) (\%) & 8.11 & 12.68 & 26.09 & 18.75 \\
\bottomrule
\end{tabular}
}
\end{center}
\end{table}

The results demonstrate an interesting trade-off in our layer-wise adaptation mechanism. Without adaptation, where hierarchical constraints are applied uniformly across layers, the model achieves better proof complexity ratios across three benchmarks but lower pass rates. This suggests that gradually reducing the constraint strength in deeper layers through layer-wise adaptation ($\alpha_l = 1-l/L$) helps achieve better proof success rates at the cost of slightly longer proofs. The superior pass rates across all benchmarks validate that our adaptive approach effectively enhances the model's theorem proving capabilities while maintaining reasonable proof complexity. Notably, even without layer-wise adaptation, our hierarchical attention mechanism still outperforms the baseline substantially in both pass rates and proof complexity, demonstrating the effectiveness of our basic hierarchical structure design.

\subsubsection{Hierarchical Structure Variants}
To explore the impact of different hierarchical structures, we conducted ablation experiments comparing the fine-grained structure (with all five levels $T_0-T_4$) against a coarse-grained variant where levels $T_1-T_3$ are merged into a single level. Table~\ref{tab:hierarchical_variants} shows the results with best-first search ($K = 64$).
\begin{table}[h!]
\begin{center}
\caption{Comparison of hierarchical structure variants on miniF2F and ProofNet benchmarks.}
\label{tab:hierarchical_variants}
\scalebox{0.75}{
\begin{tabular}{lcccc}
\toprule
\multirow{2}{*}{Method} & \multicolumn{2}{c}{miniF2F} & \multicolumn{2}{c}{ProofNet} \\
\cmidrule(lr){2-3} \cmidrule(lr){4-5}
& Test & Valid & Test & Valid \\
\midrule
PASS (baseline) & 29.51 & 31.56 & 13.56 & 10.17 \\
PASS (coarse-grained) & 30.33 & 32.38 & 14.12 & 11.30 \\
PASS (fine-grained) & \textbf{31.56} & \textbf{34.02} & \textbf{15.25} & \textbf{11.86} \\
\midrule
$R_{avg}$ (coarse-grained) & \textbf{0.67} & 0.82 & \textbf{0.73} & 1.26 \\
$R_{avg}$ (fine-grained) & 0.76 & \textbf{0.64} & 0.84 & \textbf{0.50} \\
$\text{Diff.}$(coarse-grained) (\%) & 8.33 & 11.43 & 26.32 & 21.43 \\
$\text{Diff.}$(fine-grained) (\%) & 8.11 & 12.68 & 26.09 & 18.75 \\
\bottomrule
\end{tabular}
}
\end{center}
\end{table}
On the miniF2F benchmark, the coarse-grained variant achieves better proof complexity ratios ($R_{avg}$) on the test set (0.67 vs 0.76), while the fine-grained structure achieves better complexity on the validation set (0.64 vs 0.82). More importantly, the fine-grained approach delivers superior pass rates across both test and validation sets. Similar patterns are observed in the ProofNet benchmark. These results demonstrate that distinguishing between different reasoning elements (case analysis, type declarations, and instances) is beneficial for overall theorem-proving performance, with mixed effects on proof complexity. The fine-grained structure provides the model with more detailed information about the relationships between different elements in the proof state, enabling more accurate reasoning and higher success rates.

\subsection{Explicit Level Tags Baseline}
To further evaluate the effectiveness of different structural representation methods, we implemented a baseline that uses explicit level tags. This experiment was designed to compare two fundamentally different approaches to representing structure: directly adding explicit tags in the input data versus implicitly guiding hierarchical information flow through attention mechanisms. In this baseline, we modified the input format to include explicit tags indicating the hierarchical level of each component:
\begin{verbatim}
{
    Input: "<context>...</context>...
    <type>...</type>...
    <instance>...</instance>...
    <goal>...</goal>...",
}
\end{verbatim}

Table~\ref{tab:explicit_tags} shows the results with best-first search ($K = 64$).
\begin{table}[h!]
\begin{center}
\caption{Comparison between baseline, explicit tags approach, and our method with best-first search ($K = 64$).}
\label{tab:explicit_tags}
\scalebox{0.75}{
\begin{tabular}{lcccc}
\toprule
\multirow{2}{*}{Method} & \multicolumn{2}{c}{miniF2F} & \multicolumn{2}{c}{ProofNet} \\
\cmidrule(lr){2-3} \cmidrule(lr){4-5}
& Test & Valid & Test & Valid \\
\midrule
PASS (baseline) & 29.51 & 31.56 & 13.56 & 10.17 \\
PASS (explicit tags) & 21.88 & 16.80 & 9.04 & 7.34 \\
PASS (ours) & \textbf{31.56} & \textbf{34.02} & \textbf{15.25} & \textbf{11.86} \\
\midrule
$R_{avg}$ (explicit tags) & 2.01 & 1.78 & 1.30 & 2.00 \\
$R_{avg}$ (ours) & \textbf{0.76} & \textbf{0.64} & \textbf{0.84} & \textbf{0.50} \\
$\text{Diff.}$(explicit tags) (\%) & 24.44 & 17.50 & 40.00 & 18.18 \\
$\text{Diff.}$(ours) (\%) & 8.11 & 12.68 & 26.09 & 18.75 \\
\bottomrule
\end{tabular}
}
\end{center}
\end{table}
The explicit tags approach significantly underperformed both the baseline and our method across all datasets. These results indicate that simply annotating data with explicit level tags is ineffective and potentially detrimental for theorem proving.

\subsection{Input Parsing Algorithm}
To implement our hierarchical structure, we developed a rule-based parsing algorithm that identifies different structural components in theorem text, as shown in Algorithm~\ref{alg:parsing}.
\begin{algorithm}[t!]
\caption{Hierarchical Structure Parsing}
\label{alg:parsing}
\KwIn{Theorem text}
\KwOut{Hierarchical level information}
Initialize empty segments list\;
\BlankLine
\tcc{Initialize context level}
Add context tokens to segments with level $\mathtt{context}$\;
Extract Lean4 code block between context\;
$\mathtt{current} \gets \mathtt{context}$\;
\BlankLine
\tcc{Process text line by line}
\For{each line in Lean4 code}{
\tcc{Determine hierarchical level}
\uIf{line contains $\mathtt{\vdash}$}{
$\mathtt{current} \gets \mathtt{goal}$\;
}
\uElseIf{line starts with $\mathtt{case}$}{
$\mathtt{current} \gets \mathtt{case}$\;
}
\uElseIf{line contains $\mathtt{Type}$ and $\mathtt{:}$}{
$\mathtt{current} \gets \mathtt{type}$\;
}
\uElseIf{line contains $\mathtt{:}$ but not $\mathtt{Type}$ or $\mathtt{\vdash}$}{
$\mathtt{current} \gets \mathtt{instance}$\;
}
\Else{
Maintain $\mathtt{current}$ for continued lines\;
}
Record segment with position and level\;
}
\Return{Hierarchical level information}\;
\end{algorithm}
This lightweight pattern-matching approach identifies key mathematical components through syntactic indicators: goal statements (containing turnstile $\vdash$), case analysis statements (starting with "case"), type declarations (containing "Type" and colon), and instance definitions (containing colon but neither "Type" nor turnstile). The parser maintains context across multi-line statements by inheriting the level of previous lines when appropriate, ensuring accurate hierarchical structure capture with minimal computational overhead.

\subsection{Case Studies}

To demonstrate the effectiveness of our hierarchical attention mechanism in generating concise proofs, we present three representative examples from different mathematical domains in the miniF2F dataset.

These examples showcase how our hierarchical attention mechanism improves proof generation across different mathematical domains. In Table~\ref{tab:case_study_1}, our model directly combines the function definition with the given value, eliminating the need for intermediate expansion. Table~\ref{tab:case_study_2} demonstrates improved pattern recognition, where our model directly applies the appropriate modular multiplication rule instead of decomposing the operation into addition. Table~\ref{tab:case_study_3} shows enhanced tactic understanding, combining function expansion with field simplification in a single step.
The consistent reduction in proof steps across these diverse examples demonstrates how our hierarchical attention mechanism enables better mathematical reasoning.

\begin{table}[h]
\begin{center}
\fontsize{7}{10}\selectfont  
\begin{tabular}{p{0.45\textwidth}}
\hline
\textbf{Lean4 Statement} \\
\begin{alltt}
theorem mathd_algebra_148 (c : Real) (f : Real -> Real) 
  (h0 : \(\forall\) x, f x = c * x^3 - 9 * x + 3)
  (h1 : f 2 = 9) : c = 3
\end{alltt} \\
\hline
\textbf{Baseline Proof} \\
\begin{alltt}
rw [h0] at h1    -- Substitute f(2) with its definition
linarith         -- Solve the resulting equation c * 8 - 18 + 3 = 9
\end{alltt} \\
\hline
\textbf{Our Proof} \\
\begin{alltt}
linarith only [h0 2, h1]  -- Directly solve using h0 applied to 2 and h1
\end{alltt} \\
\hline
\end{tabular}
\end{center}
\caption{Case Study 1: Basic Algebra Problem}
\label{tab:case_study_1}
\end{table}

\begin{table}[h]
\begin{center}
\fontsize{7}{10}\selectfont  
\begin{tabular}{p{0.45\textwidth}}
\hline
\textbf{Lean4 Statement} \\
\begin{alltt}
theorem mathd_numbertheory_185 (n : Nat) 
  (h0 : n % 5 = 3) : 2 * n % 5 = 1
\end{alltt} \\
\hline
\textbf{Baseline Proof} \\
\begin{alltt}
rw [two_mul]           -- Convert 2 * n to n + n
rw [Nat.add_mod, h0]   -- Apply modular addition: (3 + 3) % 5 = 1
\end{alltt} \\
\hline
\textbf{Our Proof} \\
\begin{alltt}
rw [Nat.mul_mod, h0]   -- Apply modular multiplication: 2 * 3 % 5 = 1
\end{alltt} \\
\hline
\end{tabular}
\end{center}
\caption{Case Study 2: Number Theory Problem}
\label{tab:case_study_2}
\end{table}

\begin{table}[H]
\begin{center}
\fontsize{7}{10}\selectfont  
\begin{tabular}{p{0.45\textwidth}}
\hline
\textbf{Lean4 Statement} \\
\begin{alltt}
theorem amc12a_2016_p3 (f : Real -> Real -> Real)
  (h0 :  \(\forall\) x,  \(\forall\) (y) (_ : y != 0), 
        f x y = x - y * Int.floor (x / y)) :
  f (3/8) (-(2/5)) = -(1/40)
\end{alltt} \\
\hline
\textbf{Baseline Proof} \\
\begin{alltt}
simp [h0]                  -- Expand function definition
field_simp [two_ne_zero]   -- Simplify rational expressions
norm_cast                  -- Convert between types
\end{alltt} \\
\hline
\textbf{Our Proof} \\
\begin{alltt}
field_simp [h0]    -- Combine function expansion and field simplification
norm_cast          -- Convert between types
\end{alltt} \\
\hline
\end{tabular}
\end{center}
\caption{Case Study 3: Advanced Algebra Problem}
\label{tab:case_study_3}
\end{table}

\end{document}